\documentclass[letterpaper]{article} % DO NOT CHANGE THIS
\usepackage{aaai24}  % DO NOT CHANGE THIS
\usepackage{times}  % DO NOT CHANGE THIS
\usepackage{helvet}  % DO NOT CHANGE THIS
\usepackage{courier}  % DO NOT CHANGE THIS
\usepackage[hyphens]{url}  % DO NOT CHANGE THIS
\usepackage{graphicx} % DO NOT CHANGE THIS
\usepackage{natbib}  % DO NOT CHANGE THIS AND DO NOT ADD ANY OPTIONS TO IT
\usepackage{caption} % DO NOT CHANGE THIS AND DO NOT ADD ANY OPTIONS TO IT
\usepackage{algorithm, soul, kotex}
\usepackage{algorithmic}

\usepackage{newfloat}
\usepackage{listings}
\DeclareCaptionStyle{ruled}{labelfont=normalfont,labelsep=colon,strut=off} % DO NOT CHANGE THIS
\floatstyle{ruled}
\newfloat{listing}{tb}{lst}{}
\floatname{listing}{Listing}
\usepackage{multirow, comment}
\usepackage{graphicx}
\usepackage{booktabs}
\usepackage{xcolor}
\usepackage{amsmath}
\usepackage{amssymb}
\newcommand{\simon}[1]{\textcolor{red}{Simon: #1}}

\newcommand{\hy}[1]{\textcolor{blue}{hy: #1}}
\newcommand{\sy}[1]{\textcolor{cyan}{sy: #1}}
\newcommand{\hlcyan}[1]{{\sethlcolor{cyan}\hl{#1}}}

\title{Layer Attack Unlearning: Fast and Accurate Machine Unlearning via\\ Layer Level Attack and Knowledge Distillation}
\author{
    Hyunjune Kim\textsuperscript{\rm 1}, Sangyong Lee\textsuperscript{\rm 2}, Simon S. Woo\textsuperscript{\rm 1 \rm 2 \rm 3}
}

\affiliations {
    \textsuperscript{\rm 1} Department of Applied Data Science, Sungkyunkwan University, South Korea\\
    \textsuperscript{\rm 2} Department of Artificial Intelligence, Sungkyunkwan University, South Korea\\
    \textsuperscript{\rm 3}Department of Computer Science and Engineering, Sungkyunkwan University, South Korea\\
    hyunjune.kim@g.skku.edu,
    sang8961@g.skku.edu, swoo@g.skku.edu
}

\usepackage{bibentry}
\usepackage{amsmath}

\begin{document}

\maketitle

% arxiv 각주
%\footnote{The full paper with all technical appendices is on arXiv.}

%\footnote{The full paper with all technical appendices is on arXiv.}\label{note1}

\begin{abstract}
% 한글
% 잊혀질 권리는 정보 주최자가 온라인 상에서 데이터를 삭제될 것을 요구할 수 있는 권리이다. 일반적인 서비스에서 컨텐츠 프로바이더들은 데이터베이스 상에서 SQL 쿼리로 정보 주최자의 요구에 따라 데이터를 삭제했다. 하지만 AI 서비스의 수요가 급증하면서 머신러닝 모델에서 정보 주최자의 데이터를 삭제해야 하는 일이 발생할 것이다. 머신러닝 모델에서 기존 데이터를 유지하면서 정보 주최자의 데이터만 삭제한다는 것은 쉽지 않은 문제이다. 그런 어려운 문제로 인해 Machine unlearning 작업이 필요하다. Machine unlearning은 학습된 모델에서 정보 주최자의 데이터를 망각시키는 것이 주된 목표이다. 그런 목표를 실현하기 위해 우리는 학습된 모델 내 목적성에 맞는 Layer의 weigth를 미세조정하여 망각 작업을 시키고자 한다. 우리의 방식은 Forward-Forward 알고리즘와 Boundary Unlearning 영감을 받아 모델 내 Layer를 망각 작업에 포함 시켜 전체 모델에서만 수행하던 기존 unlearning 방식의 한계를 극복하고자 한다. Knowledge Distillation(KD)을 활용하여 망각된 모델이 original model의 기능을 유지하려는 시도를 해보고자 한다.

% 영어
Recently, serious concerns have been raised about the privacy issues related to training datasets in machine learning algorithms when including personal data. Various regulations in different countries, including the GDPR, grant individuals to have personal data erased, known as ‘the right to be forgotten’ or ‘the right to erasure’. However, there has been less research on effectively and practically deleting the requested personal data from the training set while not jeopardizing the overall machine learning performance. In this work, we propose a fast and novel machine unlearning paradigm at the layer level called layer attack unlearning, which is highly accurate and fast compared to existing machine unlearning algorithms. We introduce the Partial-PGD algorithm to locate the samples to forget efficiently. In addition, we only use the last layer of the model inspired by the Forward-Forward algorithm for unlearning process. Lastly, we use Knowledge Distillation (KD) to reliably learn the decision boundaries from the teacher using soft label information to improve accuracy performance. We conducted extensive experiments with SOTA machine unlearning models and demonstrated the effectiveness of our approach for accuracy and end-to-end unlearning performance.

\end{abstract}
\section{Introduction}
\label{sec:intro}

Deep neural networks (DNNs) have achieved significant progress and dramatic performance gains in challenging machine learning tasks in recent years. Among others, large amounts of available training datasets have been the foundation for enabling the revolution of large-scale models. However, recently, due to the privacy concerns raised by ChatGPT~\cite{bourtoule2021machine,article_chatgpt_2023}, the training dataset would contain personal information or possible information that can leak one's privacy. For example, many vision-based applications would involve training one's face images, which are personally identifiable information (PII). Several nations have implemented some types of regulations, such as the General Data Protection Regulation (GDPR)~\cite{mantelero2013eu} and the EU/US Copyright Law~\cite{article_ftc_2023, article_valoha_2023}, in order to address the potential misuse of personal information and further grant individuals the right to have personal data erased, known as `the right to be forgotten' or `the right to erasure.' The goal of such regulations is to provide data owners the right to request and erase the personal or copyrighted data they want if they have not agreed and consented in the first place.

Therefore, companies using personal data should delete the requested data from the training set. One potential approach for corporations to mitigate the aforementioned concerns involves the exclusion of the required dataset from the training dataset, followed by a complete retraining process from scratch. Nevertheless, as models like ChatGPT get bigger and datasets grow, retraining them from scratch requires excessive computational resources and time.
\begin{comment}
    Nevertheless, when the models' size expands and the datasets grow, retraining these models from scratch becomes arduous and highly inefficient. \hl{For example, the retraining cost of ChatGPT would require a highly excessive amount of computational resources and time, rendering it impractical for many organizations.}
\end{comment}

\begin{comment}
    To tackle this challenge, \textit{machine unlearning} has emerged, allowing ML models to discard specific data selectively.~\cite{bourtoule2021machine}
\end{comment}
\textit{Machine unlearning} has emerged to tackle this challenge, allowing ML models to discard specific data selectively.~\cite{bourtoule2021machine} Machine unlearning can be divided into two primary strategies: \textit{instance-wise} and \textit{class-wise} unlearning.
\begin{comment}
    The former involves forgetting knowledge related to a specific set of instances from ML models, while the latter entails completely removing particular classes from ML models. Our focus lies in the class-wise unlearning paradigm, where we focus on removing specific classes entirely.
     \hy{, the pre-trained models can expose sensitive personal details like religion, diseases, or nationality on social networks, which may cause discomfort.}
\end{comment}
The former involves forgetting knowledge related to specific instances from ML models, while the latter, which we focus on, completely removes particular classes from ML models. 
For example, face recognition and social media classification systems may need to erase data related to specific religion, nationality, age, disease, gender, etc., for security and privacy reasons.
\begin{comment}
    한 문장으로는 호흡이 길어보여 단축
    For example, face recognition systems and particular group classification systems on social networking sites such as religion, nationality, age, diseases, gender and more, used for security and privacy in scenarios can utilize this method when the information of specific individuals needs to be erased.
\end{comment}
A few approaches~\cite{chen2023boundary, cha2023learning} have explored the adversarial attacks for unlearning by harnessing the forgetting data's noise to navigate the adjacent latent space. However, they used the original PGD~\cite{madry2017towards} for unlearning, which can be slow.

In this work, we propose \textbf{Layer Attack Unlearning}, a fast and novel machine unlearning algorithm to tackle the class-wise unlearning problem. Our approach first introduces \textbf{Partial-PGD}, which is a new adversarial attack generation strategy to efficiently search the close vicinity of the data points to delete (See Fig.~\ref{fig:pgdvsppgd}). Our proposed Partial-PGD is designed only to attack fully connected (classification) layer for probing the neighboring latent space to shift the forgetting data. Surprisingly, we do not utilize any feature layer information while achieving efficiency and accuracy. As shown in Fig.~\ref{fig:pgdvsppgd}, Partial-PGD is much more efficient than the original PGD, as it can create adversarial examples only via the classification layer. 

In particular,~\citet{hinton2022forward}'s Forward-Forward (FF) algorithm has inspired us, and we provide the foundation of the concept of layer-based attack for machine unlearning based on FF. According to~\citet{hinton2022forward}, each layer undergoes individualized training in the Forward-Forward algorithm to achieve its specific objectives. Similarly, in line with FF research, we aim to accomplish machine unlearning objectives at layers with characteristics directly relevant to data and features we want to forget. 
Hence, we focus on performing \textbf{machine unlearning at the layer level} rather than using the entire model. Our layer-wise unlearning approach clearly avoids unnecessary loss calculations during the unlearning process. Furthermore, updating only the layers' weights related to forgetting data will ensure a reduction in computational costs.
\begin{comment}
    Finally, we utilize Knowledge Distillation (KD)~\cite{hinton2015distilling} to adjust and retain the decision boundary for data to retain and forget. 
\end{comment}

Finally, we employ Knowledge Distillation (KD)~\cite{hinton2015distilling} to modify the decision boundary for the forgetting data and preserve the decision boundary for the retain data. The primary objective in unlearning is to utilize hard labels and acquire soft label information from the teacher model for unlearning tasks to maintain performance. We show that it achieves a stable placement of forgetting data in the space subjected to carefully created adversarial examples. We incorporated KD into our final loss function to improve performance.

Our main contributions are summarized as follows:
\begin{itemize}
    \item We introduce Layer Attack Unlearning (LAU) algorithm, which is a novel and fast unlearning method by proposing Partial-PGD and performing unlearning at the layer level.   

    \item In addition, we propose KD method to further improve the overall accuracy and data erasure performance by effectively distilling the decision boundary knowledge from the teacher model for unlearning task. 

    \item Our extensive experimental results with seven baselines with four different backbones, including ViT over three other datasets, show that our approach outperforms previous SOTA methods in accuracy and time performance while completely forgetting the requested class. 
\end{itemize}

\begin{comment}
    Our extensive experimental results with seven baselines with four different backbones, including ViT over three other datasets, clearly show that our approach outperforms previous SOTA methods in accuracy, and time performance, while completely forgetting the entire requested class. 
\end{comment}

\section{Related Work} 
\label{sec:related_work}

There are two main approaches to the current machine unlearning problem in DNNs. The first involves considering unlearning during the learning process, while the second focuses on fine-tuning. This paper will refer to the approach that considers the learning process as ``data-driven" and the approach that involves fine-tuning as ``model-agnostic."

% data-driven unlearing 작업은 학습 단계에서 최대한 망각 작업을 고려해서 설계하는 방식이다. 대표적으로 SISA라는 알고리즘이 있다.
% training 기반은 학습하는 SISA, Selective Forgetting 등과 같이 학습을 unlearning을 고려해서 데이터를 학습하는 방식이다. SISA의 경우는 데이터를 샤드와 슬라이스 단위로 나누고 중간 check point를 만들어 언제든지 모델을 롤백할 수 있게 해서 unlearning작업을 수행한다. Selective Forgetting의 경우는 Lifelong Learning 프레임워크를 제안하였으며 니모닉 코드를 합성하여 unlearning을 할 수 있게 하는 방식을 제안하였다.

% data-driven는 data partitioning, 데이터 증강등 데이터 관점에서 문제를 해결하는 접근 방법이다.

% data-driven의 unlearning 알고리즘으로는 SISA와 Selective Forgetting 등과 같은 방식이 있다. SISA의 경우는 데이터를 샤드 단위로 나누고 그들을 slices단위로 나누어 순차적으로 학습하면서 여러개의 모델의 checkpoint를 만드는 방식이다. 만약 unlearning query가 들어오게 되면 query가 학습되기 전 상태의 checkpoint로 돌아가서 다시 train을 하는 방식이며 앙상블 기법을 활용한다. 하지만 data point별로 unlearning query가 들어올 확률을 계산해 놔야 한다는 어려움을 가지고 있다. 또 다른 하나는 Selective forgetting이라는 방식이다. 해당 방식은 lifelong learning 기반으로 학습을 하여 unlearning을 수행하는 방식으로  mnemonic code라는 signal을 데이터에 학습할 때 삽입해 놓는다. 그리고 unlearning 작업을 수행할 때 선택적으로 mnemonic code 정보를 loss에 포함시켜서 forget data를 지우는 방식이다. 본 방식은 모든 데이터를 mnemonic code를 저장해 놔야하는 어려움을 가지고 있다. 이와 같은 data-driven 방식은 orinal model을 만들기 전에 unlearning query을 고민하고 있다.

\subsection{Data-Driven Unlearning Methods}
A ``data-driven" approach utilizes data-centric strategies such as partitioning and augmentation~\cite{nguyen2022survey} to address unlearning. 
SISA~\cite{bourtoule2021machine} and Selective Forgetting~\cite{shibata2021learning} are two representative data-driven unlearning methods.
\begin{comment}
    A ``data-driven" unlearning approach involves addressing unlearning from a data-centric perspective, incorporating strategies such as partitioning and augmentation~\cite{nguyen2022survey}. Approaches such as SISA~\cite{bourtoule2021machine} and Selective Forgetting~\cite{shibata2021learning} are among the methods based on the data-driven unlearning framework.
\end{comment}
 In SISA, data is divided into shard units, sequentially trained in slices, and multiple model checkpoints are created. Once an unlearning query is requested, it reverts the query to the checkpoint before learning and retrains this reverted query with the ensemble technique. However, it is challenging to calculate the probability of encountering unlearning queries on data points.  

On the other hand, Selective Forgetting~\cite{shibata2021learning} involves lifelong learning to perform unlearning. A ``mnemonic code'' signal is embedded in the data during training. During the unlearning process, the mnemonic code information is selectively incorporated into the loss function to remove forgetting data. 
This strategy requires storing mnemonic codes for all data points, considering unlearning queries before building the original model. This could be more practical in a real-world scenario.

\subsection{Model-Agnostic Unlearning Methods}
A ``model-agnostic" approach is a methodology for handling the unlearning process by adjusting the model's learning parameters to achieve data unlearning~\cite{nguyen2022survey}. Such approaches include various methods such as Summation form~\cite{cao2015towards}, Negative Gradient~\cite{golatkar2020eternal}, Fisher Forgetting~\cite{golatkar2020eternal}, Boundary unlearning~\cite{chen2023boundary}, Instance-wise Unlearning~\cite{cha2023learning}, etc. Some methods utilize adversarial attacks to the original model to avoid naively excluding and deleting forgetting data. Among the mentioned algorithms, approaches like ours include Boundary unlearning and Instance-wise Unlearning. These two algorithms perform unlearning by utilizing adversarial attacks to transition forgettable data to nearby spaces. However, a significant difference between our approach and these methods lies in the target of the attack. Our approach directs the unlearning process towards layers with specific classification objectives instead of using entire layers.  
Furthermore, we aim to introduce effective ways of utilizing PGD in unlearning.

\begin{figure}[t!]    \includegraphics[width=\linewidth]{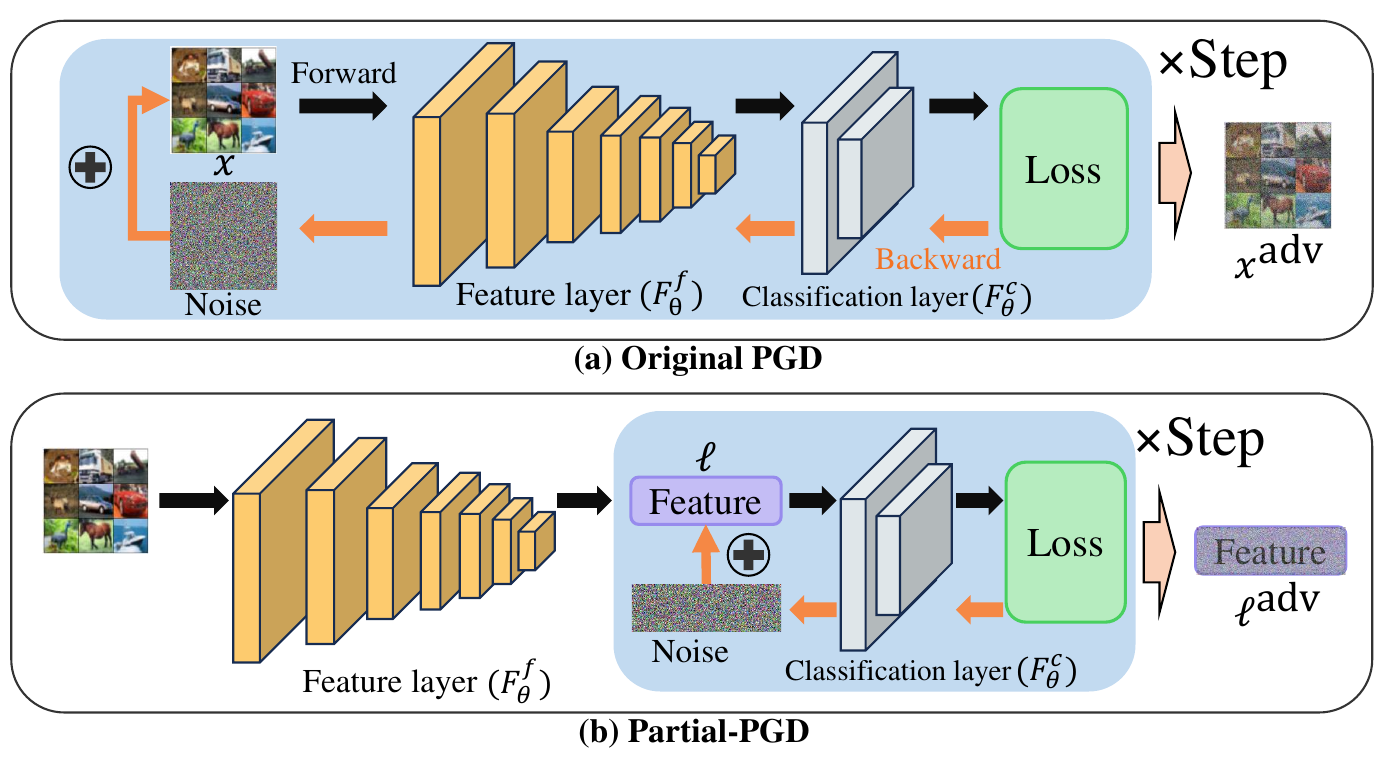}
    \caption{Illustration of the original PGD vs. Partial-PGD. While the original PGD involves backpropagation to compute $x^{\text{adv}}$ with respect to input $x$ for all the layers, (b) Partial-PGD computes $\ell^{\text{adv}}$ in $\mathcal{F}^c_{\theta}$ after passing $x$ through $\mathcal{F}^f_{\theta}$ to calculate $\ell$. Step in both (a) and (b) indicates the iteration.}
    \label{fig:pgdvsppgd}
% \vspace{-1.0em}
\end{figure}

\begin{figure*}[t] 
\includegraphics[width=\textwidth]{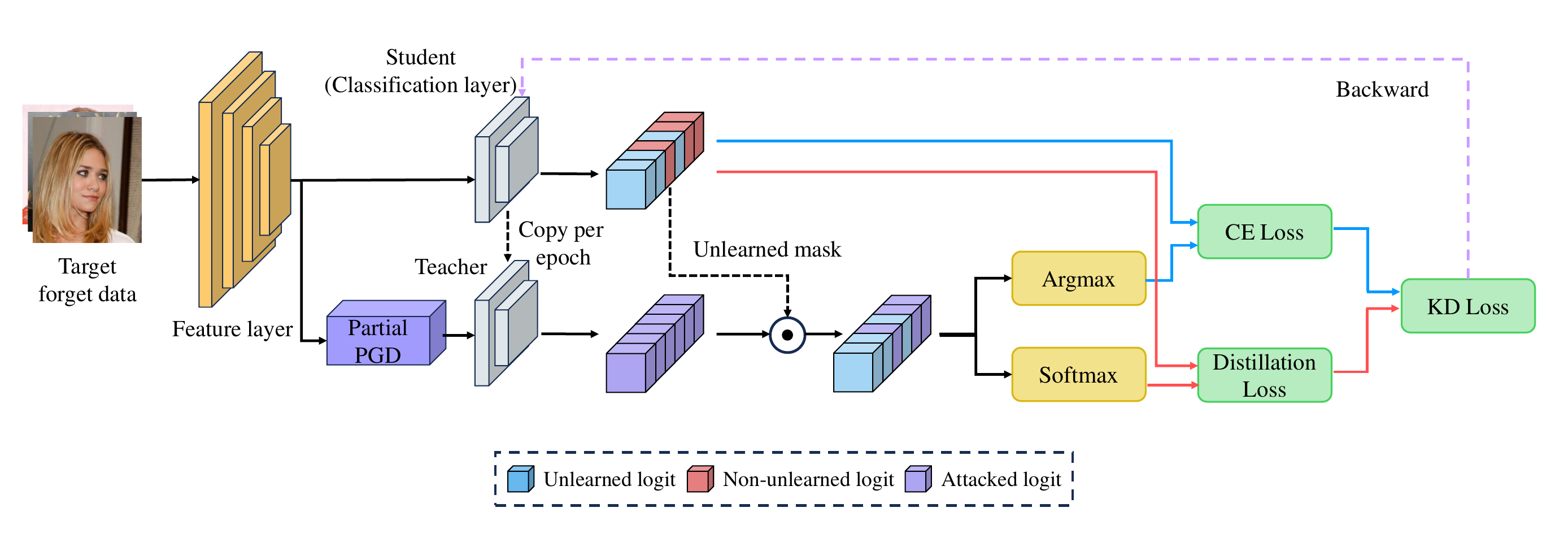}
    \caption{The overall procedure of our approach. Our method involves the unlearning task on the classification layer instead of the entire model, where each classification layer represents the student and the teacher model. For the unlearning task, we perform Knowledge Distillation by combining the teacher logit and student logit via the unlearned mask. The teacher logit is derived from the adversarial examples obtained after applying Partial-PGD.}
    \label{fig:overall}
\end{figure*}

\section{Our Approach}\label{sec:approach}

% 우리 알고리즘의 목표는 분류기에서 특정 class를 모델에서 삭제하는 class-wise unlearning이다. original model의 기능을 유지시키기 위하 지식 증류 기법을 활용하고 있다. 알고리즘 설계에 있어서 "Summation form" 논문에서 언급한 "timeliness"와 "completeness"에도 초점을 맞춰서 알고리즘을 설계했다. "timeliness"측면에서는 forgetting data set만을 사용과 partial unlearning을 수행하여 전체 모델의 loss를 계산하지 않음으로써 이득을 취하고자 하였고 "completeness" 측면에서는 forget class를 완전히 모델에서 삭제하고 정확도를 유지하려고 노력했다.

% 1) Partial-PGD는 PGD와는 달리 부분적으로 PGD 연산을 수행하여 forgetting data의 근처 공간을 찾아주며 연산 수행의 이득을 가져다 준다. 2) Classification Layer unlearning은 문제의 목적성이 뚜렸한 CNN모델의 마지막 layer들을 공격하여 unlearning을 수행한다. 3) 지식 증류와 Partial-PGD attack의 협동으로 forgetting data를 안전하게 unlearning process를 수행한다. 

The main objective of our approach is to accurately and efficiently perform \textit{class-wise} unlearning, which is to completely remove specific classes from the classification model. In this section, we describe our Partial-PGD, KD architecture, and our connection to the FF algorithm.

\subsection{Preliminaries and Notations}
First, we formulate a machine unlearning problem as follows:
We define a training dataset $D_{\text{train}}$ = ${\{{x}^{i}, {y}^{i}\}}_{i=1}^{N}$, consisting of inputs ${x}^{i} \in X$ and their corresponding class labels ${y}^{i} \in Y$. The forgetting dataset $D_{f}$ is a subset of $D_{\text{train}}$ that we intend to forget from the pre-trained model. Conversely, the retain dataset $D_{r} = D_{\text{train}} \setminus D_{f}$ is the dataset we want to preserve the overall performance.
% within the model to maintain the overall accuracy.

Next, we define the original model $\mathcal{M}_{\theta}:\mathbb{R}^n\rightarrow \mathbb{R}^n$, which comprises a set of feature layers denoted by $\mathcal{F}^f_{\theta}:\mathbb{R}^n \rightarrow \mathbb{R}^n$ and a fully connected layer denoted as $\mathcal{F}^c_{\theta}:\mathbb{R}^n \rightarrow \mathbb{R}^n$, where $\theta$ represents the optimal parameters for the model trained on $D_{\text{train}}$. The following provides a compositional representation of the model $\mathcal{M}_{\theta}$ as $\mathcal{F}^c_{\theta} \circ \mathcal{F}^f_{\theta}$.
Also, we denote $x^{\text{adv}}$ to represent the adversarial examples~\cite{goodfellow2014explaining} for the input data $x$. In particular, we define $\ell^{\text{adv}}$ as the adversarial example from Partial-PGD, generated from the intermediate latent feature $\ell$ obtained from the outputs of $\mathcal{F}^f_{\theta}$, as shown in Fig.~\ref{fig:pgdvsppgd}.

% Pratial PGD. 우리의 작업에서 적대적 공격 예시에 대한 목적성을 명확히 하고자 한다. 우리의 방법에서 적대적 공격 예시를 사용하는 이유는 forgetting data point를 unlearning 시키기 위해 근처 공간을 찾는 것에 목적성을 가지고 있다. 그러므로 전체 모델의 gradient를 계산해서 적대적 공격 예시를 도출하는 것은 불필요한 노이즈 발생 및 unlearning 속도를 저하 시킬 필요는 없다는 것이다. 그렇기 때문에 우리는 PGD를 개량해서 fully connected 층까지 계산된 적대적 공격 예시를 사용하여 unlearning 작업을 수행하려고 한다. 우리는 해당 방식을 Partial-PGD라고 부르며 Fig.3에 표현하겠다.

\subsection{Partial-Projected Gradient Descent~(PGD)}
The main reason for employing adversarial examples is to search and identify neighboring candidate spaces more effectively that will assign the forgetting data samples.
Assigning forgetting classes to random or irrelevant classes can dramatically reduce downstream task performance.
\begin{comment}
    For example, assigning forgetting classes to random or irrelevant classes can result in significantly lowering overall performance for the original downstream task.   
\end{comment}

%lead to the intended loss of information from $D_f$, while also potentially disrupting the boundary of $D_r$, which is not our intended outcome. 
Therefore, carefully exploring the neighboring space allows us not only to forget $D_f$ but also to preserve the decision boundary of other classes. Hence, adversarial attacks~\cite{madry2017towards,chen2023boundary} can be explored below: 

\begin{equation}
% \small
\label{eq:pgd}
x^{t+1} = \Pi(x^t + (\epsilon \cdot sign (\nabla_{x}\mathcal{L}({x}, {y}, \theta))),
\end{equation}
where the parameter $\theta$ represents the weights of the target model under attack, and generated noise for crafting adversarial examples is produced by computing the gradient $\nabla_x\mathcal{L}$ of the loss function $\mathcal{L}$ with respect to the input $x$.
This noise is added to $x^t$ and then projected using the projection method $\Pi$ to calculate $x^{t + 1}$, which is repeated ${t}$ times. Once $x^{t + 1}$ represents $x^{\text{adv}}$, it is an adversarial example.
\begin{comment}
    We repeatedly inject this noise into the input $x$ and subsequently employ the projection method $\Pi$ to project it. After iterating this procedure ${t}$ times, we obtain the result, which is known as the adversarial example $x^{\text{adv}}$.
\end{comment}

% 노이즈를 x^t에 더한 후 projected method $\Pi$에 projection하여 x^(t + 1)이 계산하며 그 값이 $x^{\text{adv}}$가 된다. 여기서 t는 반복횟수 이다.
However, we clarify the purpose of adversarial examples used in our work, which differs from prior approaches. The
original PGD approach may generate excessive noise and slow the unlearning process considerably. Therefore, there is no need to calculate gradients throughout the entire model to create adversarial examples.
\begin{comment}
However, we clarify the purpose of adversarial examples used in our work, which differs from prior approaches.
The original PGD approach may generate excessive noise and considerably slow the unlearning process. Therefore, we believe there is no need to calculate gradients throughout the entire model to create adversarial examples. 
\end{comment}

\begin{comment}
    Hence, we propose the Partial-PGD using $\mathcal{F}^c_{\theta}$ and generating adversarial examples from $\mathcal{F}^c_{\theta}$ for the unlearning process, as shown in~Fig.\ref{fig:pgdvsppgd}. The resulting Partial-PGD is an approach for effectively identifying the neighbor space to assign $D_f$ for forgetting data, similar to the conventional PGD, while it significantly reducing unlearning time, because of not using feature layer information. We define the Partial-PGD as follows:  
\end{comment}
Hence, our proposed Partial-PGD utilizes $\mathcal{F}^c_{\theta}$ to generate adversarial examples for the unlearning process, as shown in Fig. \ref{fig:pgdvsppgd}. This technique effectively identifies the neighboring space to allocate $D_f$, the forgetting data, similar to conventional PGD. However, it significantly reduces unlearning time by omitting feature layer information, as depicted in Fig.~\ref{fig:pgdvsppgd}. We define our Partial-PGD as follows:

% 앞에서 Partial-PGD를 이용하여 forgetting data와 가장 가까운 공간을 찾게 된다. 잊혀지고자 하는 공간은 data point의 위치에서 가장 가까운 공간이다. 기존 방식의 적대적 공격 예시를 계산할 때 (1) 수식을 사용하게 된다. w_0는 attack 대상 모델의 weight이며 입력 x에 대한 \nabla x라는 noise를 만든다.

% 하지만 우리의 방식은 (2) 수식과 같이 forgetting data point의 입력 x를 사용하여 근처 공간을 찾지 않고 {Layer_{feature}}에서 inference한 {x_feature}을 이용하여 적대적 공격을 수행하게 된다. ${x_feature}$ 통해 마지막 Layer인 ${Layer_{feature}}$에서 PGD을 수행하여 ${{x}}_{feature}^{adv}$ 값을 얻어와 ${D_f}$를 근처 공간으로 사상시킨다. 여기서 w_0는 attack 대상 Layer인 ${Layer_{feature}}$의  weight이며 입력 ${x_feature}$에 대한 \nabla ${x_feature}$라는 noise를 만든다.

\begin{equation}
% \small
\label{eq:partial_pgd}
\ell^{t+1} = \Pi(\ell^t + (\epsilon\ \cdot sign (\nabla_{\ell}\mathcal{L}(\ell, {y}, \theta))),
\end{equation}
\begin{comment}
    where Partial-PGD applies an adversarial attack to the intermediate latent $\ell$ obtained from $\mathcal{F}^{f}_{\theta}$, where $\ell$ undergoes gradient computation based solely on passing through $\mathcal{F}^{c}_{\theta}$, and then noise is added to $\ell^t$, and further projected using $\Pi$. Similarly, it can iteratively calculate $\ell^{t+1}$, which is repeated ${t}$ times. 
  \sy{original pgd와 partial pgd의 계산에 겹치는 내용을 줄이기?, original과 다르다는 것을 강조하기 위해 그냥 두기?}
\end{comment}
where Partial-PGD applies an adversarial attack to the intermediate latent $\ell$ obtained from $\mathcal{F}^{f}_{\theta}$, where $\ell$ undergoes gradient computation based solely on passing through $\mathcal{F}^{c}_{\theta}$.
Then, the result is mapped to the nearby space of a different label of $\ell$ and becomes $\ell^{\text{adv}}$, which we use for unlearning ${D_f}$ as knowledge to be forgotten.
% \simon{???unlearning or next step? softmax 무엇이 들어가면 맞나요?}
% 우리가 지우려는 ${D_f}$를 잊혀질 공간으로 보내기 위한 logit으로 사용한다.

% noise를 $\ell^t$ 더한 후 projected 함수 \Pi에 투영한다. 이로인해 $\ell^{t+1}$이 계산되며 t번 반복한다. 그 결과가 $\ell$의 근처 공간으로 사상 가능한 $\ell^{\text{adv}}$ 이다. 

%수식 (3)은 전체 모델에서 Adverserial Attack example의 출력 결과와 부분적 Adverserial Attack example의 출력값이 유사함을 표현하고 있다.

% partial unlearning 과정 그림 추가

% Partial unlearning이란 풀고자 하는 문제에 목적성이 뚜렷한 layer들 만을 가지고 unlearning을 수행하는 것이다. 우리의 목표는 class-wise unlearning문제를 푸는 것이다. 

% Forward-Forward 알고리즘은 각 layer는 자신만의 objective function을 가지고 있고 positive data에 대해서 high goodness와 negative data에 대해서는 low goodness를 충족한다. 이와 같이 우리가 풀고자 하는 class-wise unlearning 문제는 분류기에서 특정 class를 제거하는 것이 목표이다. 이런 문제와 목적성이 유사한 Layer는 CNN모델 내 ${Layer_{feature}}$이다. 우리는 ${Layer_{feature}}$대해 D_f의 "goodness"를 "badness"라는 목적성으로 변경하는 것이 Partial unlearning의 아이디어이다.

\subsection{Layer Unlearning}
% forward-forward 알고리즘은 직접적으로 unlearning과는 연관성이 없다. 하지만 앞에서 설명했 듯 모델의 각 Layer들은 목적성을 지니고 있다. 우리는 그런 각 layer들의 목적성에서 영감을 얻었다. 그런 점에서 특정 layer의 목적성을 unlearning 작업에 맞게 조심스럽게 변경해보려고 한다. CNN에서 보듯 크게 feature를 추출하는 feature layer와 classification문제를 푸는 fully connected layer 층 이렇게 2개의 파트로 구성돼있다. 이처럼 특정 목적에 맞는 layer만 unlearning 작업에 투입하여 망각 작업을 수행한다면 update해야하는 학습 파라미터 수도 줄어 모델에 부담을 줄일 수 있을 것으로 보여진다. 우리의 연구에서는 이와 같은 방식으로 분류 문제를 푸는 작업이기 때문에 목적성이 맞는 분류에 목적성을 가지고 있는 fully connected layer에 집중하고자 한다. 

While other approaches use entire layers for unlearning, we focus on unlearning only the relevant layers.
Inspired by the FF technique, we focus on the classification layer $\mathcal{F}^c_{\theta}$ to forget specific classes in the model for class-wise unlearning.
\begin{comment}
    Inspired by the FF algorithm, for the class-wise unlearning problem, we aim to forget particular classes from the model by mainly focusing on the classification layer $\mathcal{F}^c_{\theta}$ responsible for classification.
\end{comment}
 Therefore, our layer unlearning focuses on only modifying the parameters of ${\mathcal{F}^c_{\theta}}$ tied to classification instead of the entire layers and model $\mathcal{M}_{\theta}$ to forget ${D_f}$ effectively.

% 수식 5에서는 unlearned model을 표현했다. 우리의 방식을 unlearning 작업에서 $Layer_{fc}$ 층을 공격하여 ${D_{f}}$를 모델에서 제거할 것이다. 수식이 결과가 Pratial unlearning의 결과이다. 우리는 분류 문제를 푸는 것이 목적이며 목적성에 맞는 모델 내 {Layer_fc}만을 unlearning 시켜 목표를 달성하고자 한다. 목적성있는 Layer의 weight만 update함으로써 불필요한 weight의 update를 피함으로써 unlearning process 속도를 올리기 위함이다. 추가적인 $Layer_{fc}$의 unlearning process에 대해서는 아래에서 설명하고자 한다.

% Eq.\ref{eq:unlearned_model} represents the outcome of Layer Unlearning. 

We define the following equation to describe our unlearning process, where we focus on the $\mathcal{F}^c_{\theta}$ during the unlearning process to remove ${D_{f}}$ from the model:

\begin{equation}
\small
\label{eq:unlearned_model}
     \mathcal{M}_{\theta^*} = \mathcal{F}_{\theta^*}^{c} \circ \mathcal{F}^f_{\theta},
\end{equation}
where $\theta^*$ is the ideal parameters after forgetting $D_f$.
\begin{comment}
    where $\theta^*$ is the optimal parameters for unlearning after completely forgetting $D_f$.
\end{comment}

% Our objective is to solve a classification problem, and we aim to achieve our goal by unlearning only the ${\mathcal{F}^c_{\theta}}$ within the model, aligning with our purpose. 
We show that layer unlearning accelerates the unlearning process by selectively updating relevant layer weights and optimizing efficiency. 
Interestingly, it outperforms models with whole layers in accuracy.
\begin{comment}
    Furthermore, surprisingly, it achieves the higher overall accuracy compared to the models that use entire layers.
\end{comment}

% 해당 방법에서 이전 학습 중인 모델로 부터 지식증류를 받는 방식으로 Loss function을 구성하였다. 또한 unlearning과정에서 unlearning이 완료한 data point에 대해서는 skip을 하는 과정을 추가하였다. 공간이 축소하는 과정에서 이미 의사결정 경계를 넘은 forget set에 대해서는 unlearning작업이 불필요 하기 때문에 skip을 수행하게 된다. 또한 지식 증류 기법을 사용하는데 있어서 Partial unlearning과 Full unlearning 2개 과정에서 teacher 모델의 soft label의 증류를 할 때 차이를 두었다. Partial unlearning 시에 unlearning을 수행할 때 미세조정을 위해서 teacher모델의 출력을 softmax을 2번 수행하여 조금 더 안정적인 soft label을 만들어서 unlearning을 수행하였다.

% (a)에서와 같이 original model을 기준으로 boundary에 대한 지식을 증류받게 된다. epoch가 진행되면서 boundary 정보는 변경이 이루어 지게 되고 그림과 같이 (b), (c) 에서 지식을 증류받게 된다.

% Fig.x에서와 같이 매 epoch마다 Teacher로 부터 지식을 전달 받으면서 $D_{f}$의 decision boundary를 줄어나간다.

\subsection{End-to-End Unlearning Process}

% 여기에서는 Figure2.에서 표현된 unlearning process에 대해 설명하고자 한다. unlearning process는 목적성을 지닌 레이어를 unlearning하기 위한 작업이다. 또한 그 target은 분류에 특화되있는 ${Layer_{fc}$이다. Figure2.에서 나와있듯 original model에 target인 ${Layer_{fc}$ 을 복사해오게 된다. 목표는 ${Layer_{fc}$가 goodness인 $D_{f}$를 badness로 인식시켜 unlearning 작업을 완료하는 것이다. 자세한 unlearing task과정은 Figure3.에 도식화 하였다. Figure 3.에 표현했듯이 처음은 original의 ${Layer_{fc}}$가 Teacher로 사용하고 매번 epoch가 종료되면 이전 작업의 student의 ${Layer_{fc}}$가 Teacher로 변신하게 된다. 그렇게 하면서 점진적으로 $D_{f}$의 decision boundary를 안정적으로 좁혀나가게 된다.

We describe our end-to-end unlearning process, where we apply the KD to improve the overall performance further. As illustrated in Fig.~\ref{fig:overall}, the classification layer ${\mathcal{F}^c_{\theta}}$ serves as our \textit{student} model $\mathcal{S}_{\theta}$. Additionally, at the beginning of each epoch, we duplicate the $\mathcal{S}_{\theta}$ as our \textit{teacher} $\mathcal{T}_{\theta}$.
The model uses forgetting data $D_f$ as input to create an intermediate latent feature $\ell_f$ through the feature layer $\mathcal{F}_{\theta}^f$.
\begin{comment}
    If the forgetting data $D_f$ is fed into the model as an input, it passes through the feature layer $\mathcal{F}_{\theta}^f$ to create an intermediate latent feature $\ell_f$.
\end{comment}
 Then, $\ell_f $ becomes an adversarial example $\ell^{\text{adv}}_f$ after applying a Partial-PGD on the $\mathcal{T}_{\theta}$. 

Next, $\ell_f$ and $\ell^{\text{adv}}_f$ are passed through $\mathcal{S}_{\theta}$ and $\mathcal{T}_{\theta}$, respectively, becoming logits for each student and teacher, as shown in Fig.~\ref{fig:overall}. Then, the logit obtained from $\mathcal{S}_{\theta}$ is compared with the ground truth $y_f^{}$. If a discrepancy is observed, it is considered unlearned. Then, the unlearned logit replaces the adversarial logit from $\mathcal{T}_{\theta}$. This student's logit is used to compute the cross-entropy loss as follows:

% 우리의 unlearning process는 retain class에 대해서 의사결정 경계를 유지하면서 forget class의 영역을 제거하는 작업이다. unlearning을 수행하면서 최대한 retain class의 의사격정 경계를 유지하기 위해 지식 증류를 Loss function으로 활용하였다. $y^{'}$ != $y$처럼 unlearning이 완료과 된 data point에 대해서는 epoch를 수행하면서 skip하였다. 그렇게 함으로써 unlearning 속도를 올리고 모델에 부담을 줄여주기 위함이다.   
\begin{equation}
% \small
\label{eq:ce_loss}
\mathcal{L}_{CE} =
\begin{cases}
  \text CE(\mathcal{S}_{\theta}(\ell_f) , y_f^{\text{adv}}) & \text{if } y_{\mathcal{S}_{\theta}}^{} = y_f^{} \\
  \text CE(\mathcal{S}_{\theta}(\ell_f) ,  y_{\mathcal{S}_{\theta}}^{})   & \text{otherwise},
\end{cases} 
\end{equation}
where $y_{\mathcal{S}_{\theta}}^{}$ represents the predicted label from $\mathcal{S}_{\theta}(\ell_f)$, and CE is the cross-entropy function. This loss leaves the unlearned data in a state, where it makes wrong (unlearned) predictions. If not, it is trained to be a predicted label $y_f^{\text{adv}}$ of adversarial logit, leading to its unlearning process.
Next, let $Z$ be the double Softmax representation, which is defined as:
% Eq.\ref{eq:ce_loss}에서 $s_{\theta}$ 는 student를 inference한다는 의미이며 2가지 경우로 나누어 Loss를 받게 된다. $s_{\theta}(x_f) = y_f$ 경우는 $D_{f}$가 아직 unlearned한 상태가 아니며 teacher의${Layer_{fc}}$를 PGD attack으로 계산한 $x_{f}$의 근처공간인 y_f^{adv}의 정보를 받게 된다. $s_{\theta}(x_f) \neq y_f$는 $D_f$가 unlearned된 상태이며 자기 자신의 상태를 정보로 증류 받고 공간상 제자리를 지키게 된다.

\begin{equation}
% \small
\label{eq:double_softmax}
  \textnormal{Z} =
  \begin{cases}
    \sigma(\mathcal{T}_{\theta}(\ell_f^{\text{adv}})) & \text{if }y_{\mathcal{S}_{\theta}}^{} = y_f^{} \\
    \sigma(\mathcal{S}_{\theta}(\ell_f))   & \text{otherwise},
  \end{cases}
\end{equation}
where $\sigma$ represents Softmax function.
% Eq.7 에서 우리는 teacher 모델의 출력의 확률 분포를 조절하기 위해 softmax를 2번 수행하여 지식을 증류하였다. 그렇게 함으로써 student 모델로 soft label information을 전달하고자 하는 의도를 가지고 있다. 이로인해 우리가 $Layer_{fc}$만을 unlearning할 때 class간 경계를 강화하고 데이터 셋에서 낫은 정확도를 확보할 수 있었다. double softmax가 없이 Layer기반 unlearning에서 Fashion-MNIST 데이터 셋등 특정 class에서 불안한 정확도를 보였다. 해당 부분에 대해서는 Table. 과 같이 실험해 보았다. $t_{\theta}$ 는 teacher를 inference한다는 의미이며 2가지 경우로 Loss를 계산한다. 수식 \ref{eq:ce_loss}와 마찬가지로 unlearned가 아직 안 된경우와 된 경우로 나눈다.
In Eq.~\ref{eq:double_softmax}, we performed double Softmax to distill knowledge by adjusting the probability distribution of the output from $\mathcal{T}_{\theta}$. This approach is intended to convey soft label information to $\mathcal{S}_{\theta}$.
Exclusively unlearning $\mathcal{F}^c_{\theta}$ maintains the decision boundaries of retain data, and slightly improves the overall accuracy. But, layer unlearning without double Softmax showed variable accuracy, as shown in the Fashion-MNIST dataset~\cite{xiao2017fashion}.
\begin{comment}
    As a result, by exclusively unlearning $\mathcal{F}^c_{\theta}$, we can maintain boundaries between retain classes and achieve a slightly improved accuracy in the dataset. Indeed, layer unlearning without double Softmax exhibited inconsistent accuracy in specific classes, such as within the Fashion-MNIST~\cite{xiao2017fashion} dataset. 
\end{comment}
We show this effect in Section~\ref{abl}.
% Regarding this, we conducted experiments in Table.~\ref{tab:softmax_vs_double_softmax}. 
%Similar to Eq. \ref{eq:ce_loss}, \ref{eq:double_softmax} is divided into cases where unlearning has not yet occurred and where it has occurred \simon{unclear}.
Next, we define our distillation loss as follows:

% Eq. 8에서 teacher model의 soft label에서 지식을 증류 받게 된다. 이때 student 모델의 출력과 teacher 모델의 출력간 Loss를 계산하여 teacher 모델과 유사한 형태를 만들기 위해 집중하게 된다. teacher model을 도움으로 정확도를 확보하면서 Df를 모델에서 지우는데 도움을 줄 것이다. 온도는 하이퍼 파라미터로 온도를 올리면 더 부드러운 soft label을 생성하여 student모델이 teacher모델을 모방할 수 있게 도움을 준다.
\begin{equation}
% \small
\label{eq:kl_loss}
    \mathcal{L}_{DI} = \text{KL}\left(\sigma(\cfrac{\mathcal{S}_{\theta}(\ell_f)}{T}), \sigma(\cfrac{\textnormal{Z}}{T})\right),
\end{equation}
\begin{comment}
    where knowledge is distilled from $\textnormal{Z}$ of $\mathcal{T}_{\theta}$ \hl{and {\text KL} is the KL-divergence}. During distillation, the loss, $\mathcal{L}_{DI}$, between the outputs of $\mathcal{S}_{\theta}$ and $\mathcal{T}_{\theta}$ is computed to focus on creating a similar boundary to the teacher model. Additionally, $\mathcal{T}_{\theta}$ will assist in ensuring performance, while removing $D_f$ from the model, with the temperature \hl{T} as a hyper-parameter. Generally, increasing $T$ will generate smoother soft labels that assists $\mathcal{S}_{\theta}$ in mimicking $\mathcal{T}_{\theta}$.
\end{comment}
where knowledge is distilled from $\textnormal{Z}$ of $\mathcal{T}_{\theta}$ and KL is the KL divergence. During distillation, the computation of loss $\mathcal{L}_{DI}$ between the outputs of $\mathcal{S}_{\theta}$ and $\mathcal{T}_{\theta}$ focuses on creating a similar boundary to the teacher model, ensuring performance while removing information of $D_f$. The temperature $T$ is a hyper-parameter. Generally, increasing $T$ will generate smoother soft labels that assists $\mathcal{S}_{\theta}$ in mimicking $\mathcal{T}_{\theta}$. The effects of changes in $T$ are described in Suppl. Mat.

Using $\mathcal{L}_{CE}$ and $\mathcal{L}_{DI}$, our final loss function is constructed as follows:

\begin{equation}
\label{eq:final_loss}
    \mathcal{L} = (1 - \alpha) \cdot \mathcal{L}_{CE} + \alpha \cdot T^{2} \cdot \mathcal{L}_{DI},
\end{equation}
where the value of $\alpha$ represents the weight assigned to the loss between $\mathcal{L}_{CE}$ and $\mathcal{L}_{DI}$. 
As a hyper-parameter, $\alpha$ ranges from 0 to 1. Assigning additional weight to $\mathcal{L}_{CE}$ may boost unlearning time but decrease performance.
\begin{comment}
The value of $\alpha$ is also a hyper-parameter and lies within the range of 0 to 1. If we assign more weight to $\mathcal{L}_{CE}$, the unlearning speed may increase, but accuracy could decrease.
\end{comment}
  Conversely, if we provide more weight to $\mathcal{L}_{DI}$, the unlearning speed may slow down but can increase accuracy. We conducted the ablation study for $\alpha$ values to capture the trade-off. The effects of changes in the exponent of $T^{2}$ are described in Suppl. Mat.
%Finding the appropriate values for the unlearning process would require aligning suitable values with the dataset and the layer.

\begin{algorithm}[t!]
\small
\caption{End-to-End Unlearning Process}
\label{alg:unlearning_task}
\textbf{Input}: $\mathcal{F}^f_{\theta}$, $\mathcal{F}^c_{\theta}$, $D_{f}$\\
\textbf{Parameter}: Learning rate $\eta$, Hyper-parameters $\alpha$, Temperature ${T}$, Number of Epochs ${E}$\\
\textbf{Output}: {$\mathcal{M}_{\theta^*}$}
\begin{algorithmic}[1] %[1] enables line numbers
\STATE $\mathcal{S}_{\theta} \leftarrow \mathcal{F}^c_{\theta}$
\STATE $\theta^* \leftarrow \theta$
\FOR{$e$ in range $E$}
  \STATE $\mathcal{T}_{\theta^*} \leftarrow \mathcal{S}_{\theta^*}$
    \STATE ${\mathcal{L}} \leftarrow (1 - \alpha) \cdot {\mathcal{L}}_{CE} + \alpha \cdot T^2 \cdot \mathcal{L}_{DI}$

    \STATE $\theta^* \leftarrow \theta^* - \eta \cdot \mathcal{L}$
  \IF {$\mathcal{F}^c_{\theta^*} \circ \mathcal{F}^f(X_f) != Y_f$}
    \STATE \textbf{break}
  \ENDIF
\ENDFOR
\STATE $\mathcal{M}_{\theta^*} \leftarrow \mathcal{F}^c_{\theta^*} \circ \mathcal{F}^f_{\theta}$
\STATE \textbf{return} $\mathcal{M}_{\theta^*}$
\end{algorithmic}
\end{algorithm}

\begin{figure}[t!]    \includegraphics[width=0.95\linewidth]{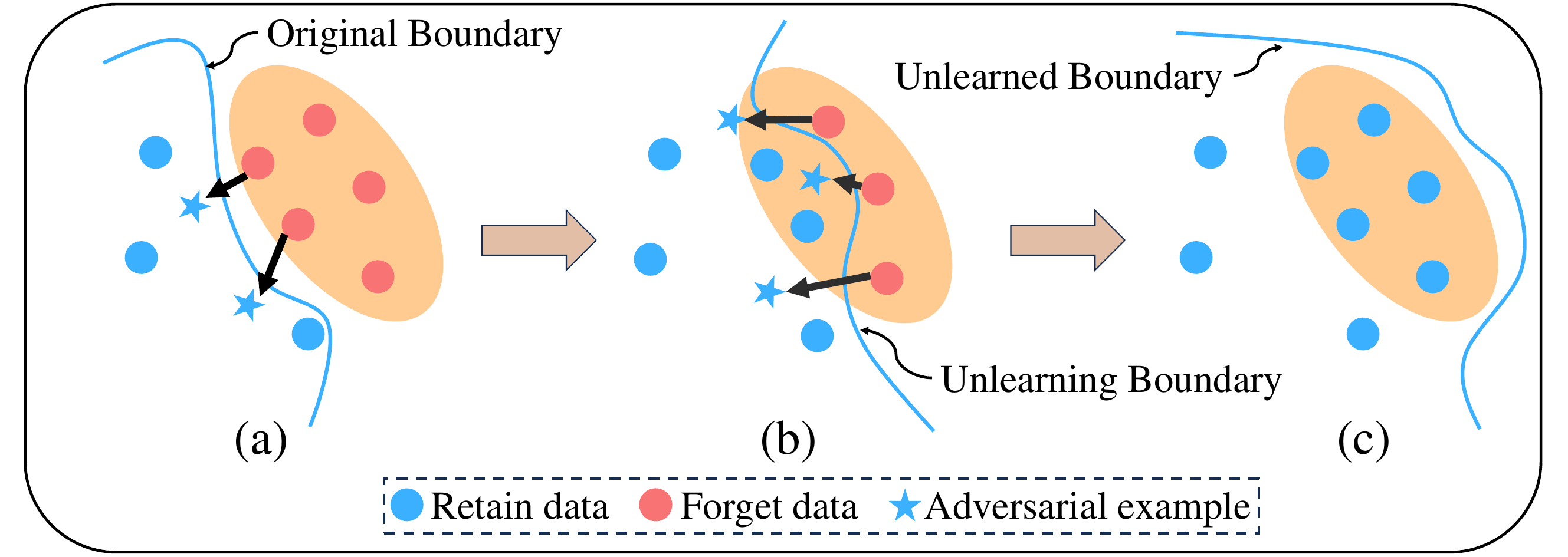}
    \caption{Boundary evolution in the unlearning process. As shown in (a), the original model receives the initial knowledge about the boundary. As the epoch progresses, the boundary information updates as depicted in (b) and (c) from the distilled knowledge.}
    \label{fig:unlearning_task}
\end{figure}

% Alg.1 에 unlearning process과정을 슈도 코드로 표현했다. unlearning 작업이 시작되기 전 지식 증류를 위해 $Layer_{teacher}$를 복사해 온다. 모든 epoch가 종료 되거나 epoch 중 batch내 $D_{f}$가 unlearned된다면 작업을 종료한다.

In addition, we provide the end-to-end unlearning process in Alg.~\ref{alg:unlearning_task}. %Before commencing the unlearning process, $\mathcal{T}_{\theta}$ is duplicated for knowledge propagation. 
We distill knowledge from $\mathcal{T}_{\theta}$, while gradually reducing boundaries. Algorithm~\ref{alg:unlearning_task} finishes either when all epochs are completed or when $D_{f}$ becomes unlearned within a batch during an epoch. Finally, we obtain our unlearning model $\mathcal{M}_{\theta^*}$ by combining $\mathcal{F}^f_{\theta}$ with the classification layer, $\mathcal{F}^c_{\theta^*}$, as shown in Eq.~\ref{eq:unlearned_model}.

\begin{comment}
    \simon{The circled area indicates the original $D_f$} \simon{스타:적대적공격을 통해 찾은 근처 공간, 그쪽으로 가기를 희망하지만, For convenience, we visualize the change from the boundary perspective, as boudary changes.}
\end{comment}
% Fig.3와 같이 우리의 방식은 매 epoch 마다 이전의 teacher로 부터 지식을 증류 받고 Partial-PGD를 통해 ${D_f}$를 모델에서 제거해 나가는 것이 Unlearning process를 수행한다.

%\begin{figure}[t!]    \includegraphics[width=\linewidth]{main/figure/kd.pdf}
%    \caption{Boundary evolution in the unlearning process. As shown in (a), the original model receives the initial knowledge about the boundary. As the epoch progresses, the boundary information updates as depicted in (b) and (c) from the distilled knowledge.}
%    \label{fig:unlearning_task}
%\end{figure}

\textbf{Summary. } In Fig.~\ref{fig:unlearning_task}, we pictorially describe our end-to-end unlearning process by displaying the boundary change for the retain and forgetting data. 

%our approach involves receiving knowledge from the previous teacher at each epoch and utilizing Partial-PGD to progressively remove ${D_f}$ from the model, thereby performing the unlearning process.

\section{Experimental Results}\label{sec:experiments}

% 한글로 우선 작성, 실험 모델, 데이터 추가(vggface2), 2차원 평면 공간 이미지 준비
% boundary unlearning 실험을 참고
% Resnet또는 VGG

% 아래와 같이 표 만들기
% data set cifar 10, cifar100
% cifar 10 테스트 모델 list
  % AllCNN
  % VGG16
  % Resnet18
  % Resnet50
% cifar100
  % Resnet50

%Dr, Df, Drt, Dft 비교표
% 비교 모델 (ALLCNN, VGG16, Resnet16, Resnet50)
% Original, 경사상승, Boundary unlearning, Part unlearning, FPU, FPU + MAS

% t-SNE 데이터 시각화 추가 (공간 표시 및 grid table 생성)

% 엑셀 파일 작성
\begin{table*}[htp!]
\centering
\caption{Accuracy and Unlearning Score (US) performance on the CIFAR-10, Fashion-MNIST and VGGFace2 datasets. %~\cite{krizhevsky2009learning, xiao2017fashion, cao2018vggface2}. 
Bold font highlights the highest performing results, while underlining indicates the second-best performance.}
\label{tab:accuracy_table}
\resizebox{\textwidth}{!}{%
\begin{tabular}{c|c|cccc|c|cccc|c|cccc|c|cccc|c}
\toprule
                               & Model             & \multicolumn{5}{c}{VGG16}                                                                                                           & \multicolumn{5}{c}{ResNet18}                                                                                                        & \multicolumn{5}{c}{ResNet50}                                                                                                        & \multicolumn{5}{c}{ViT}                                                                                                             \\ \cline{2-22}
                               & Metrics            & $D_{r} \uparrow$ & $D_{f} \downarrow$ & $D_{tr} \uparrow$ & $D_{tf} \downarrow$ & US & $D_{r} \uparrow$ & $D_{f} \downarrow$ & $D_{tr} \uparrow$ & $D_{tf} \downarrow$ & US & $D_{r} \uparrow$ & $D_{f} \downarrow$ & $D_{tr} \uparrow$  & $D_{tf} \downarrow$ & US & $D_{r} \uparrow$ & $D_{f} \downarrow$ & $D_{tr} \uparrow$ & $D_{tf} \downarrow$ & US \\ \midrule \midrule
\parbox[t]{2mm}{\multirow{9}{*}{\rotatebox[origin=c]{90}{CIFAR-10}}}      & Original          &99.98                                &100                                  &92.07                               &96.70                  &0.4494
               &99.98                               &100                                  &93.13                               &96.60              &0.4575
                   &99.94                                &99.96                                  &93.44                               &95.0      &0.4646                          &88.06                                &93.52                                  &81.48                               &88.40                   &0.4020              \\
                               & Retrain (Optimal) & 99.89                          & 0                                & 91.98                         & 0              &0.9390
                 & 99.79                          & 0                                & 92.50                          & 0                 &0.9428
              & 99.77                          & 0                                & 92.48                         & 0               &0.9426
                & 95.0                             & 0                                & 81.0                            & 0                 &0.8631
              \\
                               & Negative Gradient                & 88.53                          & 16.96                            & 79.86                         & 17.0       &0.7320
                       & 93.85                          & 28.38                            & 86.30                         & 25.54         &0.7204
                  & 88.75                          & 24.77                            & 82.52                         & 23.30          &0.7087
                  & 85.264                         & 18.69                            & 79.74                         & 16.7              &0.7332
              \\
                               & Fine-tune         & 99.63                          & 0                                & 90.09                         & 0              &\underline{0.9253}
                 & 99.63                          & 0                                & 91.25                         & 0              &\underline{0.9337}
                 & 99.45                          & 0                                & 90.79                         & 0              &\underline{0.9304}
                 & 90.96                          & 1.77                             & 82.43                         & 1.62               &\underline{0.8598}
             \\
                               & Random Label      & 80.99                          & 3.56                             & 72.40                          & 3.69            &0.7805                & 91.38      
                    & 11.09                            & 84.00                         & 10.98          &0.8007
                 & 81.30                          & 12.91                            & 76.62                         & 11.84             &0.7467
              & 77.58                          & 15.10                            & 73.42                         & 14.38                 &0.7094
          \\
                               & Fisher Forgetting & 46.78                          & 55.24                            & 44.61                         & 52.30           &0.3414
                 & 59.0                             & 52.34                            & 55.57                         & 52.2           &0.3945
                 & 58.17                          & 58.06                            & 55.95                         & 56.20            &0.3781
                & 42.68                            & 66.34                              & 43.34                           & 62.30           &0.2911
                  \\
                               & Boundary Shrink   & 90.73                         & 10.16                            & 81.53                         & 9.58             &0.7943
               & 95.88                          & 9.75                             & 87.91                         & 10.24            &0.8329
               & 86.03                          & 3.94                             & 80.09                         & 3.46             &0.8303
               & 85.22                          & 0.61                             & 79.29                         & 0.28             &0.8498
               \\
                               & IWU               & 90.81                          & 0                                & 82.35                         & 0.10             &0.8712
                & 89.41                          & 0                                & 82.55                         & 0               &0.8733
                & 86.11                          & 0                                & 79.98                         & 0               &0.8564
                & 82.48                          & 3.92                             & 77.01                         & 2.58            &0.8173
                \\ 
                               & \textbf{Ours}               & 99.97                          & 0                                & 92.18                         & 0    &\textbf{0.9405}                         & 99.97                          & 0                                & 93.53                         & 0       &\textbf{0.9504}                        & 99.92                          & 0                                & 93.52                         & 0            &\textbf{0.9503}                   & 87.51                          & 0                             & 81.14                         & 0         &\textbf{0.8640}                   \\ \midrule
\parbox[t]{2mm}{\multirow{9}{*}{\rotatebox[origin=c]{90}{Fashion-MNIST}}} & Original          &99.83                                &100                                  &94.38                               &99.60          &0.4579                       &98.45                                &99.96                                  &94.71                               &99.70              &0.4601                   &98.49                                &99.98                                  &94.68                               &99.6               &0.4601                  &91.27                                &98.71                                  &88.28                               &97.10                &0.4210                 \\
                               & Retrain (Optimal) & 100                            & 0                                & 93.40                          & 0            &0.9494
                   & 100                            & 0                                & 93.38                         & 0            &0.9493
                   & 100                            & 0                                & 93.28                         & 0            &0.9485
                   & 89.44                          & 0                                & 86.76                         & 0            &0.9019                   \\
                               & Negative Gradient                & 97.77                          & 0                                & 92.63                         & 0              &0.9438
                 & 92.57                          & 1.39                             & 90.04                         & 0.84           &0.9183
                 & 84.44                          & 12.63                            & 81.42                         & 10.22          &0.7890
                 & 71.77                          & 0.10                              & 70.38                         & 0.10          &0.7964                   \\
                               & Fine-tune         & 99.67                          & 0                                & 93.07                         & 0               &0.9470
                & 97.23                          & 0                                & 91.93                         & 0            &\underline{0.9386}
                   & 98.83                          & 0                                & 92.85                         & 0         &\underline{0.9454}
                      & 96.08                          & 0.01                            & 88.72                         & 0.10    &\textbf{0.9148}                         \\
                               & Random Label      & 98.17                          & 8.34                             & 92.43                         & 23.55              &0.7763
             & 76.80                          & 11.47                            & 74.80                          & 11.54          &0.7375
                 & 75.99                          & 10.77                            & 73.73                         & 10.72       &0.7368
                    & 84.18                          & 11.36                            & 82.10                         & 13.04    &0.7736                       \\
                               & Fisher Forgetting & 62.33                          & 28.81                            & 60.32                         & 28.10         &0.5471
                   & 72.78                          & 57.65                            & 71.03                         & 54.10         &0.4705
                   & 60.59                          & 84.01                            & 60.25                         & 82.60         &0.2958
                   & 43.42                           & 88.01                             & 42.60                          & 86.3          &0.1972
                  \\
                               & Boundary Shrink   & 86.88                         & 1.47                             & 81.66                        & 1.12            &0.8586
                & 95.78                          & 34.54                            & 92.31                         & 32.40             &0.7225               & 83.50                         & 30.23                            & 80.60                          & 27.08             &0.6728              & 70.31                          & 2.04                             & 68.74                         & 2.70               &0.7665          \\ 
                               & IWU               & 99.09                          & 0                                & 93.68                         & 0            &\underline{0.9515}                   & 93.82                          & 0                                & 90.80                         & 0              &0.9304                 & 80.17                          & 0                                & 77.94                         & 0               &0.8434                & 82.85                          & 0                                & 81.21                         & 0             &0.8645                  \\  & \textbf{Ours}               & 99.51                          & 0                                & 93.89                         & 0         &\textbf{0.9531}                      & 97.98                          & 0                                & 94.54                         & 0            &\textbf{0.9579}                   & 98.14                          & 0                                & 94.48                         & 0          &\textbf{0.9575}                    & 90.11                          & 0                                & 87.44                         & 0                 &\underline{0.9066}              \\ \midrule
\parbox[t]{2mm}{\multirow{9}{*}{\rotatebox[origin=c]{90}{VGGFace2}}}      & Original          &100                                &100                                  &96.67                               &98.41     &0.4787                            &100                                &100                                  &95.88                               &98.41              &0.4727                   &99.12                                &98.43                                  &93.67                               &100              &0.4514                   &94.71                                &96.86                                  &95.43                               &93.82  &0.4832                               \\
                               & Retrain (Optimal) & 99.98                          & 0                                & 96.67                         & 0              &0.9740                 & 100                            & 0                                & 96.20                         & 0               &0.9705                & 99.10                          & 0                                & 94.77                         & 0              &0.9596                 & 92.63                          & 0                                & 93.32                         & 0             &0.9488                  \\
                               & Negative Gradient                & 96.85                          & 15.67                            & 90.50                         & 4.76        &0.8915                    & 97.32                          & 9.75                             & 89.55                         & 12.69           &0.8272                & 86.80                          & 4.73                             & 78.79                         & 3.17            &0.8241                & 91.16                          & 1.63                             & 92.34                         & 0                   &\underline{0.9416}            \\
                               & Fine-tune         & 97.86                          & 0                                & 89.87                         & 0              &\underline{0.9416}                 & 91.42                          & 0                                & 85.91                         & 0               &0.8960                & 95.18                          & 0                                & 90.03                         & 0              &\underline{0.9249}                 & 96.91                          & 1.63                             & 84.85                         & 3.70               &0.8600             \\
                               & Random Label      & 90.32                          & 1.74                             & 79.11                         & 1.58           &0.8384                 & 96.76                          & 6.44                             & 87.34                         & 0                  &\underline{0.9059}             & 88.24                          & 13.19                            & 82.43                         & 9.52               &0.8007             & 92.06                          & 9.68                             & 91.04                         & 8.64               &0.8667             \\
                               & Fisher Forgetting & 46.24                          & 31.01                            & 42.72                         & 50.79           &0.3400                & 72.78                          & 57.65                            & 71.03                         & 54.10                &0.4705            & 76.28                          & 4.52                             & 71.83                         & 7.93            &0.7455                & 60.80                           & 71.07                             & 53.58                          & 60.49              &0.3472             \\
                               & Boundary Shrink   & 99.48                          & 17.25                            & 93.04                         & 5.36              &0.9055              & 94.02                          & 5.40                              & 86.08                         & 5.36          &0.8559                  & 93.85                          & 5.36                            & 85.78                         & 5.0            &0.8565                   & 86.92                          & 6.46                                & 86.81                         & 4.25              &0.8693
                 \\
                               & IWU               & 99.21                          & 10.80                            & 94.46                         & 4.76                &0.8650            & 75.23                          & 0.17                             & 69.77                         & 0            &0.7936                   & 78.62                          & 0                                & 69.14                         & 0           &0.7899                    & 76.25                          & 0.27                             & 78.66                         & 0                   &0.8479           \\ 
                               & \textbf{Ours}               & 99.70                          & 0                            & 96.70                         & 0        &\textbf{0.9743}                    & 99.79                          & 0                             & 95.34                         & 0         &\textbf{0.9639}                      & 97.46                          & 0                                & 93.28                         & 0                  &\textbf{0.9485}             & 95.18                          & 0                             & 95.50                         & 0             &\textbf{0.9651}                 \\ \bottomrule
\end{tabular}%
}
% \vspace{-1.0em}
\end{table*}
We experiment and evaluate popular unlearning benchmarks used in other unlearning research~\cite{golatkar2020eternal, chen2023boundary, cha2023learning} on image classification tasks.

\noindent \textbf{Datasets and Models. } We conducted experiments on CIFAR-10~\cite{krizhevsky2009learning}, Fashion-MNIST~\cite{xiao2017fashion}, and VGGFace2~\cite{cao2018vggface2} datasets. For the VGGFace2 dataset, we randomly select ten individuals from a training dataset containing over 600 images, ensuring a balanced gender distribution.
Furthermore, we perform training from scratch for three different architectures: VGG16~\cite{simonyan2014very}, ResNets~\cite{he2016deep}, and ViT~\cite{dosovitskiy2020transformers}. 

\noindent \textbf{Baseline Approaches.} The subsequent unlearning baseline methods are used:

\noindent 1) \textbf{Original}: We train the model on the $D_{\text{train}}$ dataset before undergoing the unlearning process.

% unlearning?
\noindent 2) \textbf{Retrain}: We train the model from scratch utilizing $D_{r}$ as the retrained model, an optimal unlearning strategy.

\noindent 3) \textbf{Negative Gradient (NG)}~\cite{golatkar2020eternal}: We fine-tune the \textbf{Original} with $D_{f}$ by following the direction of gradient ascent.

\noindent 4) \textbf{Fine-tune}~\cite{golatkar2020eternal}: We fine-tune the \textbf{Original} using $D_{r}$ with a large learning rate.

\noindent 5) \textbf{Random Label}~\cite{golatkar2020eternal}: We fine-tune the \textbf{Original} by assigning arbitrary labels randomly to $D_{f}$. 

\noindent 6) \textbf{Fisher Forgetting}~\cite{golatkar2020eternal}: The Fisher Forgetting model identifies influential parameters significantly affecting $D_f$ and then introduces noise to neutralize their impact.

\noindent 7) \textbf{Boundary Shrink}~\cite{chen2023boundary}: We create adversarial examples from $D_f$ and assign new adversarial labels to shrink towards different classes.

\noindent 8) \textbf{IWU}~\cite{cha2023learning}: Generating adversarial instances for distinct labels via $D_f$ and incorporating a regularization term. 
While initially designed for instance-wise learning, we adapt this method for class-wise unlearning problems.

\noindent \textbf{Implementation Details and Evaluation Metrics.} Our method and other baselines are implemented in Python 3.7 and use the PyTorch library~\cite{paszke2019pytorch}, employing a single NVIDIA GeForce RTX 3090 GPU. The initial model was trained using an LR scheduler and an SGD optimizer with specific settings (momentum: 0.9, weight decay: 5 × $10^{-4}$, initial learning rate: 0.01). For the unlearning phase, we employ the SGD optimizer and conduct experiments with varying learning rates (ranging from 0.001 to 0.01), KD $\alpha$ values (ranging from 0.3 to 0.7), KD temperature $T$ value (fixed at 4), and Partial-PGD values (ranging from 0.4 to 1.0). 
% \hl{$D_{f}$ and $D_{r}$ represent the forgetting and retain data, while $D_{tf}$ and $D_{tr}$ correspond to the test forgetting and retain data, respectively.}
As defined, $D_{f}$ and $D_{r}$ represent the forgetting and retain data, respectively. Additionally, $D_{tf}$ corresponds to the test forgetting data, and $D_{tr}$ represents the test retain data. 
We assess the accuracy of all four different datasets.

% unlearning task를 완료 하면 unlearned model은 D_f에 대해서 어떤 정보도 갖고 있으면 안된다. 그러므로 Retrain과 유사한 Acc_Dr^test와 Acc_Df^test를 확보하는 것이 우리의 목표이다. Table1.에 분류 모델, 데이터셋, Metric 별로 테스트를 진행하였다. 테스트에 사용한 모델들은 VGG16, ResNet18, ResNet50 그리고 ViT이다. 데이터셋은 CIFAR-10, Fashion-MNIST을 사용하여 테스트를 진행하였다. 또한 US(unlearning socre)를 통해 Table1.에서 가장 최고의 성능을 평가하였다. The equation is as follows:

\subsection{Accuracy Performance}
%Upon accomplishing the unlearning process, the unlearned model must be devoid of any information related to $D_{f}$. 
To achieve the best unlearning performance, it should completely forget information related to $D_{f}$. Therefore, guaranteeing accuracy on a par with those achieved by the \textbf{Retrain} for both $D_f$ and $D_r$ will be the best. Table~\ref{tab:accuracy_table} presents test results from different classification models, datasets, and metrics. The tested models include VGG16, ResNet18, ResNet50, and ViT. The datasets used for testing were CIFAR-10, Fashion-MNIST, and VGGFace2. In addition to the accuracy metric, we evaluate the performance using the unlearning score (US) as follows:
\begin{equation}
% \scriptsize
\label{eq:ue_2}
    {\text{US}(\text{acc}_r,\text{acc}_f)} = \cfrac{\exp(\cfrac{\text{acc}_r}{100})+\exp(1-\cfrac{\text{acc}_f}{100})-2}{2 \cdot (\exp(1)-1)},
\end{equation}
where $\text{acc}_r$ and~$\text{acc}_f$ denote the accuracy of the retain and forgetting dataset, respectively. If the $D_{tr}$ approaches 100\% and $D_{tf}$ approaches 0\%, the US metric approaches 1, indicating a stable result on the unlearning process. We provide a more detailed explanation of why this metric is useful for unlearning in Suppl. Mat.
%\footnote{The full version is available at ArXiv.}

% 값이 $D_{tr}$은 100%에 근사하고 $D_{tf}$는 0%에 근사하면 US가 1에 근접하게 되면 unlearning process가 안정적으로 이루어 졌다는 수치이다.
% 수식 Eq.11 로 US(unlearning score)를 계산하였다. 1에 가까우면 가장 좋은 unlearning metrics이라는 지표이다. 우리는 해당 지표를 $D_{tr}$과 $D_{tf}$ 에 적용하여 평가하였다.

% Table. 1에서 CIRAR-10, Fashion-MNIST그리고 VGGFace2에서 특정 라벨 1개에 대해서 unlearning process를 수행하고 Dr, Df, Dtr, Dtf 그리고 US를 측정하였다.

% Table 1.에서 NG의 경우 US수치로 Fashion-MNIST의 VGG16과 ResNet18, 그리고 VGGFace2의 ViT를 제외하고 0.9 이하를 기록하였다. Fine-ture의 성능은 CIFAR-10의 ViT, VGGFace2 의 ResNet18과 ViT를 제외하고 0.9 이상이라는 좋은 기록을 보여주었지만 Fashion-MNIST의 ViT에서 좋은 점수를 받았지만 D_f와 D_tf에서 0.01%가 unlearned되지 않은 것을 볼 수 있었다. Random Label에서는 전체 테스트에서 US가 대략 0.7894를 기록하였으며 하지만 VGGFace2의 ResNet18에서는 0.9059받았고 D_tf를 0%가 된 것을 볼 수 있었다. Fisher Forgetting은 전체 테스트에서 US가 0.4015라는 안 좋은 성능 지표를 보여주었다. Boundary Shrink의 경우 전체 테스트 평균 0.8175 라는 점수를 받았으며 Fashion-MNIST에서는 0.6728라는 성능을 보여주었다. IWU는 전체 테스트에서 평균적으로 US가 0.8587이라 점수를 기록하였으며 특히 Fashion-MNIST의 VGG16에서 0.9515라는 좋은 성적을 거두었다. 마지막으로 우리의 방식은 전체 테스트에서 평균적으로 0.9443이라는 좋은 점수가 나왔다. CIFAR-10의 viT에서 0.8640이라는 점수가 나왔지만 original 모델의 D_r과 D_tf와 비교했을 때 합리적은 수준이다.전체 테스트에서는 D_f와 D_tf를 완전히 제거한 것을 볼 수 있다.

% nagative gradient의 loss에 유동성이 있어서 성능적 측면에서 좋지 않은 결과를 보여주는 듯 하다.

Finally, Table~\ref{tab:accuracy_table} presents the performance of each unlearning method for a specific single class in the aforementioned datasets. We measure the accuracy for datasets $D_{r}$, $D_{f}$, $D_{tr}$, and $D_{tf}$, along with the metric US. 
For the \textbf{NG}, the unstable variability in the loss of negative gradient contributes to less favorable overall performance results.
% Fine-tuning의 경우에는 unlearning의 completeness 면에서는 우수하지만 Table.2 timeliness면에서 떨어지는 것을 볼 수 있다.
\textbf{Fine-tune} shows strong performance in forgetting and retaining information. Nevertheless, this methodology requires utilizing the complete dataset $D_r$ during training. Such extensive data is time-consuming, and we analyze and compare their worse time performance in Table~\ref{tab:time}.
% Random Label의 경우 %{D_{f}}%에 대해서 무작위 성으로 잊게하는 방식이기에 분류 공간상에느 random label로 수렴 시키기는 쉽지 않아 좋은 성능이 않나온다.
In the case of \textbf{Random Label}, except for VGGFace2's ResNet18, most cases have poor accuracy and US. Due to the random nature of forgetting, converging towards arbitrary labels in the classification space is challenging, resulting in low performance.

%US score측면에서도 좋지 않은 결과가 나왔으며 timeliness측면에서도 fisher의 matrix information을 계산해야하기 때문에 많은 시간이 소요됐다.
\textbf{Fisher Forgetting} exhibits poor performance, with low accuracy and US on the overall test. Also, the Fisher matrix information required a significant amount of time.
% 우리와 마찬가지로 적대적 공격 예시를 이용하는 방법이지만 단순히 ${D_{f}}$에 대한 공격 예시의 hard label information을 이용하여 불안정한 unlearning process를 보여줬다. 
For \textbf{Boundary Shrink}, they also utilized adversarial attack examples, but they used the hard label information of the attack examples on ${D_{f}}$, which resulted in an unstable unlearning process.
% 적대적 공격 예시를 사용하며 regularization 으로 안정적인 unlearning process를 수행하는 방식이다.
\textbf{IWU} approach involves utilizing adversarial attack examples while incorporating regularization to achieve a stable unlearning process. However, this gains an average US of 0.8587 in the overall test.
\begin{table}[t!]
\centering
% \vspace{-5mm}
\caption{Total extra data used and time consumption in $sec$ for training different unlearning methods.}
\label{tab:time}
\resizebox{\linewidth}{!}{%
\begin{tabular}{c|ccccccccc}
\toprule
\multirow{2}{*}&                               & \multirow{2}{*}{Retrain} & Fisher & Fine- & \multirow{2}{*}{NG} & Random & Boundary & \multirow{2}{*}{IWU} & \multirow{2}{*}{\textbf{Ours}} \\
&                               &  & Forgetting & tune &  & Label & Shrink &  &  \\
\midrule \midrule
\parbox[t]{2mm}{\multirow{6}{*}{\rotatebox[origin=c]{90}{CIFAR-10}}} & \textbf{Total Extra} & \multirow{2}{*}{45,000} & \multirow{2}{*}{45,000} & \multirow{2}{*}{45,000} & \multirow{2}{*}{5,000} & \multirow{2}{*}{5,000} & \multirow{2}{*}{5,000} & \multirow{2}{*}{5,000} & \multirow{2}{*}{5,000}            \\
 & \textbf{Data Used} &  &  &  &  &  &  &  &             \\ \cline{2-10}
& \textbf{Time w/ VGG16} & 3,683 & 9,710 & 433  & 73 & 24 & 116 & 1351 & \textbf{3.76} \\
& \textbf{Time w/ ResNet18} & 2,871 & 12,526 & 546  & 153 & 30 & 191 & 362 & \textbf{4.37} \\
& \textbf{Time w/ ResNet50} & 4,705 & 19,850 & 1,061  & 174 & 57 & 471 & 1513 & \textbf{7.76} \\
& \textbf{Time w/ ViT} & 4,441 & 13,238 & 479  & 78 & 23 & 163 & 1563 & \textbf{25.93} \\ \midrule
\parbox[t]{2mm}{\multirow{6}{*}{\rotatebox[origin=c]{90}{Fshion-MNIST}}} & \textbf{Total Extra} & \multirow{2}{*}{54,000} & \multirow{2}{*}{54,000} & \multirow{2}{*}{54,000} & \multirow{2}{*}{6,000} & \multirow{2}{*}{6,000} & \multirow{2}{*}{6,000} & \multirow{2}{*}{6,000} & \multirow{2}{*}{6,000}            \\
 & \textbf{Data Used} &  &  &  &  &  &  &  &             \\ \cline{2-10}
& \textbf{Time w/ VGG16} & 2,309 & 8,526 & 430  & 85 & 23 & 214 & 1072 & \textbf{8.75} \\
& \textbf{Time w/ ResNet18} & 2,768 & 12,116 & 582  & 103 & 30 & 715 & 223 & \textbf{5.19} \\
& \textbf{Time w/ ResNet50} & 5,758 & 22,013 & 1,229  & 206 & 76 & 929 & 967 & \textbf{9.14} \\
& \textbf{Time w/ ViT} & 2,155 & 8,377 & 487  & 80 & 25 & 282 & 546 & \textbf{13.39} \\ \midrule
\parbox[t]{2mm}{\multirow{6}{*}{\rotatebox[origin=c]{90}{VGGFace2}}} & \textbf{Total Extra} & \multirow{2}{*}{5,726} & \multirow{2}{*}{5,726} & \multirow{2}{*}{5,726} & \multirow{2}{*}{574} & \multirow{2}{*}{574} & \multirow{2}{*}{574} & \multirow{2}{*}{574} & \multirow{2}{*}{574}      \\
& \textbf{Data Used} &  &  &  & & &  &  &       \\ \cline{2-10}
& \textbf{Time w/ VGG16} & 1,840 & 1,295 & 468  & 400 & 17 & 338 & 548 & \textbf{5.6} \\
& \textbf{Time w/ ResNet18} & 1,861 & 1,354 & 670 & 140 & 27 & 473 &1258 & \textbf{ 6.51}\\
& \textbf{Time w/ ResNet50} & 3,721 &2,597 &3,291 &484 &157 &503 &1837 & \textbf{17.77} \\
& \textbf{Time w/ ViT} & 2,155 &1,428 &665 &84 &27 &187 &783 & \textbf{ 6.74} \\ \bottomrule
\end{tabular}%
}
% \vspace{-2em}
\end{table}

Finally, \textbf{Ours} completely removes the forgetting dataset (0\% accuracy) on all the test cases and retains the highest unlearning performance. The accuracy for both $D_{f}$ and $D_{tf}$ reaches 0, while the accuracy for $D_{r}$ and $D_{tr}$ is comparable to or sometimes even higher than the \textbf{Retrain}. Also, ours demonstrates superior performance compared to almost all baseline models across various scenarios, with a high US average of 0.9443. Our approach that utilizes Partial-PGD and KD-based unlearning processes on layers with explicit objectives clearly achieves the best unlearning performance.

\begin{figure}[b!] \centering   \includegraphics[width=\linewidth]{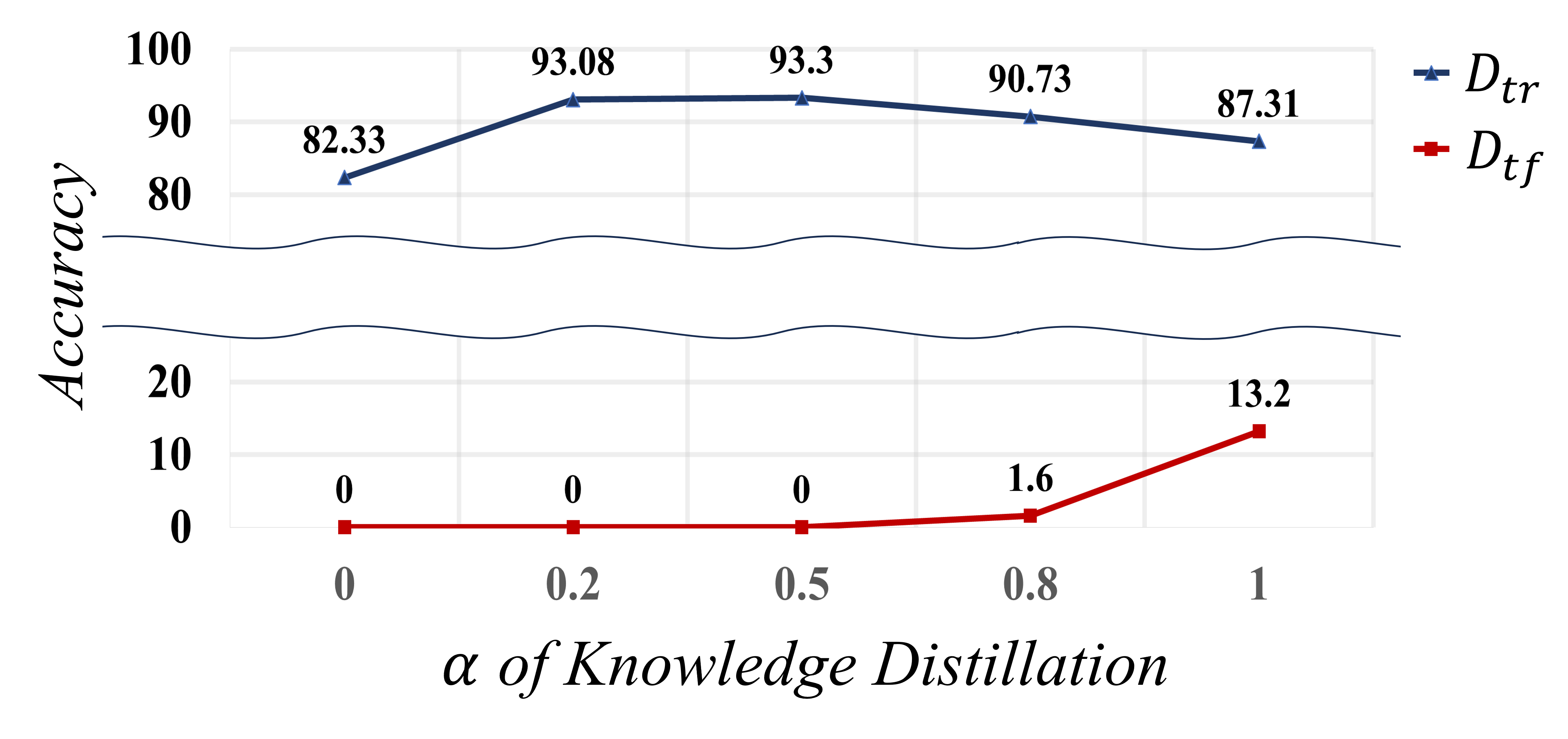}
    \caption{Impact of hyper-parameter $\alpha$ in Knowledge Distillation vs. Accuracy on CIFAR-10 with ResNet18.}
    \label{fig:multi_tsne_result}
\end{figure}

\begin{figure*}[tp!] \centering    \includegraphics[width=\textwidth]{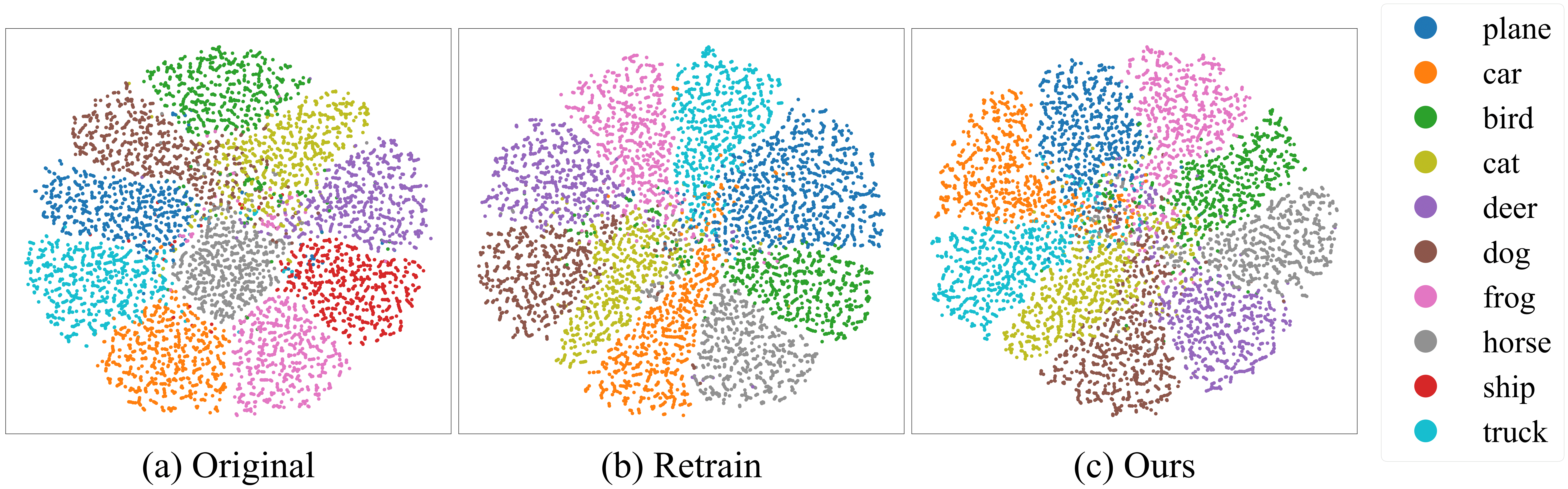}
    \caption{Visualization of decision boundary for the CIFAR-10 dataset in ResNet18, where each point represents a sample colored with the predicted classes. Red dots in (a) are the data to be removed, which are not showing in (b) and (c), indicating the successful unlearning. Similar plots for other models are provided in Suppl. Mat.}
    \label{fig:tsne_result}
\end{figure*}

% \usepackage{graphicx}
\begin{table}[t!]
\centering
\caption{Original PGD vs. Partial-PGD.}
\label{tab:pgd_vs_partial_pgd}
\resizebox{\linewidth}{!}{%
\begin{tabular}{c|ccc|ccc} \toprule
         & \multicolumn{3}{c|}{Original PGD} & \multicolumn{3}{c}{Partial PGD} \\ \cline{2-7}
         & $D_{tr}$    & $D_{tf}$  & Time~(s)    & $D_{tr}$       & $D_{tf}$     & Time~(s)      \\ \midrule \midrule
VGG16    & 92.03  & 0    & 14.18   & 92.18     & 0       & 3.76      \\
ResNet18 & 92.97  & 0    & 18.19   & 93.53     & 0       & 4.37      \\
ResNet50 & 91.84  & 0    & 44.15   & 93.52     & 0       & 7.76      \\
ViT      & 78.07  & 0    & 237.36  & 81.14     & 0       & 25.93    \\ \bottomrule
\end{tabular}%
}
\end{table}
% 우리의 방식은 Partial-PGD와 목적성이 뚜렷한 Layer만을 지식 증류기반의 unlearning process를 활용하였기에 이런 좋은 성능이 나올 수 있었다.

% Table 2에서 각 metrics 별로 elapsed time, 사용된 데이터양을 측정하였다. Retrain, Fisher Fogetting, Fin-tune에서는 D_f와 D_r 전체를 사용하였으며 NG, Random Label, Boundary Shrink, IWU, Ours에서는 D_f만을 이용하여 unlearning task를 수행하였다. 측정한 elapsed time의 단위는 초이다. 우리의 방식에서 데이터와 상관없이 VGG16의 경우 3.76초에서 8.75초가 소요됐다. ResNet18의 경우 4.37초에서 6.51초 정도가 소요됐다. ResNet50의 경우 7.76초에서 17.77이 소요됐고 ViT는 6.74초에서 25.93초가 소요됐다.

\subsection{Data Usage \& Time Performance}
Table~\ref{tab:time} presents each method's elapsed time and data usage. The \textbf{Retrain}, \textbf{Fisher Forgetting}, and \textbf{Fine-tuning} leverage the entire $D_{r}$ dataset, resulting in significant time costs for unlearning. Including our method, the rest of the unlearning methods utilize only $D_{f}$. In the case of \textbf{Fisher Forgetting}, it takes a longer time than the \textbf{Retrain}, and its unlearning performance is significantly poor. While the \textbf{Fine-tune} exhibits favorable unlearning performance, it comes with the drawback of consuming a considerable time.
\begin{comment}
    However, depending on the dataset and model, our method showcases optimal unlearning performance while consuming only 3.76 to 25.93 seconds.
    In summary, our approach demonstrates superior performance in speed and accuracy compared to alternative methods.
\end{comment}
However, our method showcases optimal unlearning performance, while consuming only 3.76 seconds in the quickest scenario. To summarize, our approach exhibits higher efficiency, compared to competing methods.

\begin{comment}
    \begin{figure}[htp!] \centering    \includegraphics[width=\linewidth]{main/figure/tsne_result1.png}
    \caption{Visualization of decision boundary for the CIFAR-10 dataset in ResNet18, where each point represents a sample colored with the predicted classes. Red dots in (a) are the data to be removed, which are not showing in (b) and (c), indicating the successful unlearning. Similar plots for other models are provided in Supplementary.}
    \label{fig:tsne_result}
\end{figure}
\end{comment}
%Table.4에서 언급한 (d)는 multi-classes를  (6,8) set 실험결과이다. 

% 해당 세션에서는 구성 요소를 시스템에서 제거하거나 변경함으로써 어떤 효과가 발생하는지에 대한 실험을 하고자 한다. CIFAR-10의 1개 class의 D_f의 양은 5000개이며 이중 각각 모델에 대해서 렌덤하게 50%, 10%를 줄여서 unlearning task를 실행시켜 보았다. {D_tr}, {D_tf}, {US} 그리고 실행시간을 측정하였다. 다음 시나리오에서 2,500개에 대해서 모든 모델이 unlearning이 완료됐지만 500개에 대해서 ViT는 0.1%가 남았다. 실행 속도는 {D_f}의 양이 줄어들면서 속도가 빨라졌다. 해당 실험에서 전체 데이터 셋을 사용하지 않고 일부의 중요한 {D_f}를 찾는 다면 보다 더 낫은 unlearning을 할 수 있는 가능성을 보여주었다.

\subsection{Ablation Study}\label{abl}
We performed several different ablation experiments to analyze and show the benefits of our approach.

\subsubsection{Original PGD vs. Partial-PGD. } Table~\ref{tab:pgd_vs_partial_pgd} compares unlearning performance when applying the original PGD vs. Partial-PGD within our method on the CIFAR-10 dataset. While the original PGD yields high unlearning performance, Partial-PGD indicates even superior outcomes. Notably, Partial-PGD accelerates the unlearning process by up to nearly tenfold compared to the original PGD.

% \usepackage{graphicx}
\begin{table}[t!]
\centering
\caption{Effect of Softmax vs. Double Softmax.}
\label{tab:softmax_vs_double_softmax}
\resizebox{\linewidth}{!}{%
\begin{tabular}{c|ccc|ccc} \toprule
         & \multicolumn{3}{c|}{w/o Double Softmax} & \multicolumn{3}{c}{w/ Double Softmax} \\ \cline{2-7}
         & $D_{tr}$      & $D_{tf}$   & Time~(s)     & $D_{tr}$                      & $D_{tf}$                    & Time~(s)                     \\ \midrule \midrule
VGG16    & 84.74    & 0     & 10.9     & 93.89                    & 0                      & 8.75                     \\
ResNet18 & 91.42    & 0.1   & 25.87    & 94.54                    & 0                      & 5.19                     \\
ResNet50 & 80.91    & 0     & 93.49    & 94.48                    & 0                      & 9.13                     \\
ViT      & 87.01    & 0     & 61.37   & 87.44                    & 0                      & 13.39               \\ \bottomrule     
\end{tabular}%
}
% \vspace{-1.0em}
\end{table}
\subsubsection{Double Softmax. } 
\begin{comment}
    Within our approach, the teacher logits are passed through passing through a softmax function before being incorporated into the distillation loss.
\end{comment}
In our technique, the teacher logits undergo a softmax function before being integrated into the distillation loss.
We have coined this method ``Double Softmax", where Double Softmax enhances the robustness of our method across diverse datasets and models. And, Table~\ref{tab:softmax_vs_double_softmax} presents unlearning performance with and without double Softmax in our methods on the Fashion-MNIST dataset.
% 지식증류의 하이퍼 파라이터 알파의 효과와 정확도

\subsubsection{Data Usage Ratio.} The class-specific $D_{f}$ dataset for one class in CIFAR-10 contains 5,000 samples. As shown in Table \ref{tab:amount_accuracy}, we reduced the dataset size to 50\% (2,500) and 10\% (500) for each model to perform the unlearning task. We measure the accuracy, US, and execution time of ${D_{tr}}$, ${D_{tf}}$. In the following scenario, all models completed the unlearning for 2,500 samples, but ViT still had 0.1\% retaining for 500 samples. The execution speed increases as the size of ${D_{f}}$ decreases. 
Our experiment shows the potential for achieving superior unlearning performance by focusing on critical subsets of ${D_{f}}$ rather than employing the complete dataset, saving time nearly seven times.

\begin{table}[t!]
\centering
\caption{The changes in time and accuracy performance with the reduction in ${D_{f}}$ data on the CIFAR-10.}
\label{tab:amount_accuracy}
\resizebox{\linewidth}{!}{%
\begin{tabular}{c|c|cc|cc|cc|cc} \toprule
                         & Model                 & \multicolumn{2}{c|}{VGG16} & \multicolumn{2}{c|}{ResNet18} & \multicolumn{2}{c|}{ResNet50} & \multicolumn{2}{c}{ViT} \\ \cline{2-10}
                         & Total Extra & \multirow{2}{*}{2,500}        & \multirow{2}{*}{500}         & \multirow{2}{*}{2,500}          & \multirow{2}{*}{500}          & \multirow{2}{*}{2,500}          & \multirow{2}{*}{500}          & \multirow{2}{*}{2,500}       & \multirow{2}{*}{500} \\ 
                         &Data Used &         &          &           &           &           &           &        &         \\ \midrule \midrule
\parbox[t]{2mm}{\multirow{4}{*}{\rotatebox[origin=c]{90}{Metrics}}}  & ${D_{tr}}$            & 92.42       & 92.38       & 93.51         & 93.38        & 93.63         & 93.37        & 81.14      & 81.6       \\
                         & ${D_{tf}}$            & 0           & 0           & 0             & 0            & 0             & 0            & 0          & 0.1        \\
                         & US                    & 0.9422      & 0.9420      & 0.9503        & 0.9493       & 0.9512        & 0.9493       & 0.8640     & 0.8662     \\
                         & Time                  & 1.91        & 1.21        & 2.28          & 1.45         & 3.81          & 1.62         & 25.63      & 14.55     \\ \bottomrule
\end{tabular}%
}
% \vspace{-1.0em}
\end{table}
\subsubsection{Hyper-parameter $\alpha$ in KD.} As shown in Fig.~\ref{fig:multi_tsne_result}, we examine the accuracy variation of ${D_{tr}}$ and ${D_{tf}}$ with respect to changes in the hyper-parameter $\alpha$ in Eq.~\ref{eq:kl_loss}.
\begin{comment}
    As the $\alpha$ value approaches 0, it strongly emphasizes the process of forgetting information from $D_f$ without taking into account any knowledge from $D_r$. 
\end{comment}
As the $\alpha$ approaches zero, it exclusively prioritizes the removal of $D_f$ without taking into account any information from $D_r$.
Consequently, the information about ${D_{tf}}$ is completely removed, resulting in a decrease in the accuracy of ${D_{tr}}$.
\begin{comment}
    As $\alpha$ approaches 1, emphasizing the retaining information through a strong reliance on the teacher model causes ${D_{tr}}$ accuracy to decrease, rendering unlearning ineffective. Therefore, selecting the appropriate $\alpha$ value is crucial in maximizing unlearning performance. Therefore, in this work, we used $\alpha$ ranging from 0.4 to 0.6.
\end{comment}
As $\alpha$ approaches one, heavily relying on the teacher model for retaining information increases ${D_{tf}}$ accuracy, indicating ineffective unlearning. Therefore, selecting the appropriate $\alpha$ value can maximize unlearning performance. Consequently, we used $\alpha$ ranging from 0.4 to 0.6 in this work. In more detail, the effects of changes in $\alpha$ are described in Suppl. Mat.

% Table4.와 같이 CIFAR-10 ResNet18에서 multi-label unlearning을 실험해 보았다. 해당 실험은 동시에 2개 class를 unlearning하는 실험이다. (2, 4), (4, 5), (6, 8) 그리고 (7, 9) 이렇게 총 4번의 multi-label unlearning task를 수행하였다. 실험에서 동일한 hyper-parameter에서 특정 label 집합을 동시에 수행할 때 unlearning 시간에 차이를 보였다. 특히 (7, 9) 라벨들에서 다른 집합의 unlearning 보다 많은 시간이 소요된 것을 볼 수 있었다.
\begin{comment}
    \input{main/table/multi_label_forget}
\subsubsection{Multi Classes Unlearning} We consider the multi-class data of the CIFAR-10 dataset as forgetting data and present the accuracy for each class after unlearning the data in Table \ref{tab:multi-label-table}. This experiment involves simultaneously unlearning two classes with the same hyper-parameters. We perform a total of four multi-class unlearning tasks: (2, 4), (4, 6), (6, 8), and (7, 9). Our method is effective not only in unlearning individual classes but also in handling multi-class scenarios, showing its ability to perform well across various class differences.
\end{comment}

% Fig 5.는 에서는 original 모델, retrain 모델 그리고 우리의 unlearned 모델을 t-sne로 표현하였다. 붉은 점은 ship image samle이며 ${D_{f}}$를 표현하고 있다. Fig 5. (b)에서 Retrain 모델에서 ${D_{f}}$가 완전히 지워졌음을 확인할 수 있다. 우리의 방식으로 unlearning task를 수행한 결과를 Fig 5. (c)에서 확인할 수 있다. ${D_{f}}$의 decision boundary가 주변 공간으로 흡수 된 것을 볼 수 있다. Fig 6.는 origianl 모델에서 우리의 방식으로 2개의 ${D_{f}}$인 ship과 frog images를 제거하여 표현된 decision boundary이다.

\subsection{Visualization on Decision Boundary}
Figure~\ref{fig:tsne_result} presents the \textbf{Original}, \textbf{Retrain}, and \textbf{Ours} using t-SNE on the CIFAR-10 dataset.
The red dots represent samples of ship images, indicated as ${D_{f}}$.
As shown in Fig.~\ref{fig:tsne_result}(b), ${D_{f}}$ was totally misclassified in the \textbf{Retrain}. On the other hand, Our unlearning method produces results correctly, as shown in Fig.~\ref{fig:tsne_result}(c), where the decision boundary of ${D_{f}}$ has been successfully absorbed into the surrounding space. 
\begin{comment}
 We visually represent \hl{the \textbf{Original}}, \hl{the \textbf{Retrain}}, and our unlearned model using t-SNE on CIFAR-10 dataset in Fig.~\ref{fig:tsne_result}. In Fig.~\ref{fig:tsne_result}~(b), it can be observed that ${D_{f}}$ has been completely removed from the \textbf{Retrain}.
The results of performing the unlearning task using our approach can be observed in Fig.~\ref{fig:tsne_result}~(c). 
\end{comment}
\section{Conclusion}\label{sec:conclusion}

% 한글로 우선 작성
% 이 논문에서 우리는 효과적인 적대적 공격 예시인 Partial-PGD와 전체 모델이 아닌 목적석이 뚜렷한 Layer만을 unlearning하는 Layer unlearning을 소개하였다. 실험에서도 알 수 있듯이 특정 Layer들만을 가지고 목적성만 변경한다면 unlearning이 가능하다는 것도 입증하였다. 또한 update하는 파라미터의 수가 전체 모델 대비 적기 때문에 소비 컴퓨티 파워도 줄여 속도적인 측면에서도 이득을 취했다. 이런 우리가 제시한 예시가 미래에는 모델 내 다른 Layer들의 목적성을 명확히 파악하여 다양한 unlearning문제를 풀 것으로 기대한다.

In this paper, we introduce a novel and fast machine unlearning algorithm, layer attack unlearning, which is the new layer-based unlearning paradigm. Our work proposes Partial-PGD, layer unlearning method, and KD end-to-end framework to improve the overall accuracy performance while completely removing the forgetting dataset.
% 수정
Through extensive experimental evaluations, we demonstrated that modifying only specific layers’ learning objectives can lead to successful unlearning.
% Our approach not only significantly reduces the number of parameters and their updates (computational cost) but also shortens the unlearning (time) performance. 
Our approach effectively decreases both the number of parameters and their updates (computational cost), consequently reducing the overall time required for unlearning.
We believe our layer attack unlearning paves a new way for future research in effectively addressing various unlearning challenges.
%We find that modifying only specific layers' learning objectives can effectively unlearn. Hence, our approach reduces the number of parameters, computational cost and time performance.
% We believe that our layer attack unlearning paves a new way for future research to address various unlearning challenges effectively.
\begin{comment}
    Through extensive experimental evaluations, we demonstrated that modifying only specific layers' learning objectives can lead to successful unlearning. Our approach not only significantly reduces the number of parameters and their updates (computational cost), but also shortens the unlearning (time) performance.
\end{comment}
\section{Acknowledgments}
The authors would thank anonymous reviewers. Simon S. Woo is the corresponding author. This work was partly supported by Institute for Information \& communication Technology Planning \& evaluation (IITP) grants funded by the Korean government MSIT: (No. 2022-0-01199, Graduate School of Convergence Security at Sungkyunkwan University), (No. 2022-0-01045, Self-directed Multi-Modal Intelligence for solving unknown, open domain problems), (No. 2022-0-00688, AI Platform to Fully Adapt and Reflect Privacy-Policy Changes), (No. 2021-0-02068, Artificial Intelligence Innovation Hub), (No. 2019-0-00421, AI Graduate School Support Program at Sungkyunkwan University), and (No. RS-2023-00230337, Advanced and Proactive AI Platform Research and Development Against Malicious deepfakes).

\bibliography{aaai24}
\clearpage
% \onecolumn
\appendix
\begin{huge}
    \begin{center}
    \textbf{Supplementary Materials}
    \label{sec:supple}
\end{center}
\end{huge}

\section{Datasets}
We used the three different datasets as follows:
% CIFAR-10
% Fashion-MNIST
% VGGFace2
\begin{itemize}
  \item \textbf{CIFAR-10.} The CIFAR-10 dataset~\cite{krizhevsky2009learning} is a widely used benchmark in classification tasks. It consists of 60,000 images in ten classes. The dataset is divided into a training set of 50,000 images and a test set of 10,000 images. We experiment to erase only one class (5,000 images) out of 10 classes.
  \item \textbf{Fashion-MNIST.} The Fashion-MNIST dataset~\cite{xiao2017fashion} is popular in classification tasks. It contains 70,000 grayscale images of various fashion items, categorized into ten classes. The dataset is divided into a training set of 60,000 images and a test set of 10,000 images. We experiment with erasing only one class (6,000 images) out of 10 classes. We utilize the dataset to evaluate unlearning performance in grayscale images.
  \item \textbf{VGGFace2.} The VGGFace2 dataset~\cite{simonyan2014very} is a large-scale face dataset designed for face recognition tasks. This dataset consists of facial data and is closely related to tasks to preserve privacy. Given the high similarity among classes, it is a crucial dataset for assessing the effectiveness of unlearning methods in real-life scenarios involving facial data. It consists of diverse face images that vary regarding identities, poses, illuminations, backgrounds, and expressions. The dataset contains over 3.31 million images from more than 9,000 individuals. But, to experiment with our unlearning task, we randomly chose ten individuals from a training dataset containing over 600 images, ensuring a balanced distribution of gender.
\end{itemize}

\section{Evaluation Metrics for Unlearning Performance}
The results in our experiments are evaluated based on the following metrics:
\subsubsection{Accuracy. }
In order to assess a classifier's performance, accuracy is frequently utilized. It measures the percentage of samples for which the true classes can be predicted with the maximum degree of certainty. Accuracy of a model $\mathcal{M}_{\theta}$ tested on a dataset of N samples $\{(x_1, y_1), ...,(x_N , y_N )\}$ is formulated as
follows:
\begin{equation}
    \text{ACC} = 100 \cdot \cfrac{\sum^N_{i=1}\delta(\sigma(\mathcal{M}_{\theta}(x_i)),y_i)}{\text{N}},
\end{equation}
where $\delta(\cdot,\cdot)$ is the Kronecker delta function.
\subsubsection{Unlearning Score (US).} 
Effective unlearning performance refers to the ability of a model to effectively forget information from the forgetting data, while concurrently retaining the relevant information from the retain data. However, determining the most effective metric for unlearning is challenging because of the orthogonal objective between evaluating and measuring forgetting vs. retain data accuracy, where they are not in a linear relationship. For instance, when comparing two unlearning approaches, one exhibits good performance in forgetting data but does poorly retain data well. At the same time, the other will be the opposite case, with poor performance in forgetting data but great accuracy in retaining data. In such scenarios, it becomes quite challenging to determine which method is better based solely on one of the two accuracies. Therefore, while considering both accuracies is essential, there is no straightforward way to assess both simultaneously. Such discrepancy leads to difficulties in evaluating unlearning performance.

Therefore, to evaluate unlearning performance more accurately and effectively, we propose and define a new metric, called Unlearning Score (US), which effectively characterizes and combines the two accuracies into a single value to assess unlearning performance. Since accuracy is measured in percentages, we normalize it to a range of 0 to 1 by dividing by 100. As accuracy for forgetting data is preferred to be lower, we subtract the value from 1 to convert it into a higher-is-better range. Next, we input the values into the exponential function. The following equation pertains to the retain data:
\begin{comment}
    \simon{explain why this metric is better than accuracy? in what aspect?
1. 이 공식이 어떻게 구성되었는지, 왜 그렇게 구성되었는지? 나누기2와 왜 하필 exp 함수를 쓰며, USf와 USr은 무엇인지, 각각 설명하시고
2. aacuracy에서 측정 못한느것을 US는 어떤것을 더 잘 측정하는지를 적으시기 바랍나다. }
\end{comment}

\begin{equation}
    \text{US}_r = \text{exp}(\cfrac{\text{acc}_{r}}{100}),
    \label{eq:us_r}
\end{equation}
where $\text{acc}_r$ is accuracy of retain data.

Similarly, the following equation is defined for the forgetting data:
\begin{equation}
    \text{US}_f = \text{exp}(1-\cfrac{\text{acc}_{f}}{100}),
    \label{eq:us_f}
\end{equation}
where $\text{acc}_f$ is accuracy of forgetting data.

In fact, we use exponential functions, which offer a good way to assign and map weights to values, much better than linear functions. In other words, rather than simply using the two accuracies as they are, this approach enables us to better characterize higher scores as accuracies increase. Conversely, for lower accuracies, we can assign lower scores.
We calculate the average of $\text{US}_r$ and $\text{US}_f$ values obtained through Eq.~\ref{eq:us_r} and Eq.~\ref{eq:us_f}, respectively. Then, we normalize them to range from 0 to 1 using $min-max$ scaling with exp(1) and exp(0).
% \usepackage{multirow}
% \usepackage{graphicx}
\begin{table*}[t]
\centering
\caption{Unlearning performance based on changes in the $\alpha$ value in knowledge distillation.}
\label{tab:total_alpha-table}
\resizebox{\textwidth}{!}{%
\begin{tabular}{c|c|ccc|ccc|ccc|ccc|ccc} \toprule
                               \multicolumn{2}{c}{$\alpha$}          & \multicolumn{3}{|c|}{0}       & \multicolumn{3}{c|}{0.2}                & \multicolumn{3}{c|}{0.5}                & \multicolumn{3}{c|}{0.8}                & \multicolumn{3}{c}{1}                  \\ \midrule
                               \multicolumn{2}{c|}{Metrics}          & $D_{tr}$ & $D_{tf}$ &US & $D_{tr}$ & $D_{tf}$&US & $D_{tr}$ & $D_{tf}$&US & $D_{tr}$ & $D_{tf}$&US & $D_{tr}$ & $D_{tf}$&US \\ \midrule
\multirow{4}{*}{CIFAR-10}      & VGG16    & 75.31    & 0   &0.8269     & 91.93    & 0    &0.9386    & 92.28    & 0   &0.9412     & 92.17    & 0    &0.9404    & 92.14    & 0     &0.9402   \\
                               & ResNet18 & 92.87    & 0    &0.9456    & 93.38    & 0    &0.9493    & 93.50    & 0    &0.9502    & 92.47    & 2.4  &0.9239    & 91.58    & 6.7   &0.8849   \\
                               & ResNet50 & 91.75    & 0    &0.9374   & 93.51    & 0     &0.9503   & 93.43    & 0    &0.9497    & 90.06    & 2.8   &0.9033   & 86       & 10.30  &0.8192  \\
                               & ViT      & 78.46    & 0    &0.8467    & 80.65    & 0    &0.8608    & 81.22    & 0   &0.8645     & 81.16    & 0    &0.8642    & 79.11    & 0.5   &0.8469   \\ \midrule
\multirow{4}{*}{Fashion-MNIST} & VGG16    & 77.38    & 0    &0.8399    & 94.21    & 0    &0.9555    & 93.91    & 0   &0.9532     & 92.78    & 0    &0.9449    & 80.47    & 4.9   &0.8218   \\
                               & ResNet18 & 91.14    & 0    &0.9329    & 93.92    & 0    &0.9533    & 94.6     & 0   &0.9584     & 93.38    & 0.6  &0.9446    & 91.54    & 7.4    &0.8794  \\
                               & ResNet50 & 85.57    & 0    &0.8937    & 94.67    & 0    &0.9590    & 93       & 0    &0.9465    & 93.18    & 0.1  &0.9471    & 92.43    & 8.3    &0.8793  \\
                               & ViT      & 88.22    & 0    &0.9121    & 88.38    & 0    &0.9132    & 88.57    & 0    &0.9146    & 88.75    & 0    &0.9158    & 88.44    & 0      &0.9136  \\ \midrule
\multirow{4}{*}{VGGFace2}      & VGG16    & 91.93    & 0    &0.9386    & 93.35    & 0    &0.9491    & 96.04    & 0   &0.9693     & 96.99    & 0    &0.9765    & 96.99    & 0     &0.9765   \\
                               & ResNet18 & 56.01    & 0    &0.7185    & 85.12    & 0    &0.8906    & 94.62    & 0   &0.9585     & 94.30    & 0    &0.9562    & 93.19    & 0      &0.9479  \\
                               & ResNet50 & 90.82    & 0    &0.9306    & 93.67    & 0    &0.9514    & 94.46    & 0   &0.9573     & 89.39    & 4.76  &0.8836   & 89.87    & 14.28  &0.8185  \\
                               & ViT      & 94.30    & 0    &0.9562    & 95.56    & 0    &0.9657    & 95.88    & 0   &0.9681     & 95.72    & 0    &0.9669    & 95.56    & 0    &0.9657   \\ \bottomrule
\end{tabular}%
}
\end{table*}
\begin{table}[t]
\centering
\caption{Unlearning performance based on changes in the $T$ value in knowledge distillation on CIFAR-10 with ResNet18.}
\label{tab:KD_T}
\resizebox{0.9\columnwidth}{!}{%
\begin{tabular}{c|cccc|c}
\toprule
T       & 1     & 4     & 8     & 16     &Original \\
\midrule \midrule
$D_{r}$  & 99.98 & 99.98 & 99.97 & 99.97 &99.98 \\
$D_{f}$  & 0     & 0     & 0     & 0     &100 \\
$D_{tr}$ & 93.4  & 93.53 & 93.32 & 93.24 &93.13 \\
$D_{tf}$ & 0     & 0     & 0     & 0     &96.60 \\
US       & 0.9495     & 0.9504     & 0.9489     & 0.9483     &0.4575 \\
\bottomrule
\end{tabular}%
}
\end{table}
\begin{table}[t]
\centering
\caption{Unlearning performance based on changes in the $x$ value of $T^{x}$ in knowledge distillation on CIFAR-10 with ResNet18.}
\label{tab:KD_2T}
\resizebox{0.9\columnwidth}{!}{%
\begin{tabular}{c|cccc|c}
\toprule
$x$       & 1       & 2     & 3     & 4        &Original \\
\midrule \midrule
$D_{r}$  & 99.97  & 99.98 & 99.95 & 90.63    &99.98 \\
$D_{f}$  & 0      & 0     & 1.94  & 4.77     &100 \\
$D_{tr}$ & 93.37  & 93.53 & 92.67 & 82.09    &93.13 \\
$D_{tf}$ & 0      & 0     & 2.7   & 5.03     &96.60 \\
US       & 0.9493 & 0.9504& 0.9230& 0.8315   &0.4575 \\
\bottomrule
\end{tabular}%
}
\end{table}
\begin{table}[t]
    \centering
    \caption{Original PGD vs. Partial-PGD for all datasets.}
    \label{tab:sub_pgd}
    \resizebox{\columnwidth}{!}{%
    \begin{tabular}{c|c|cccc|cccc}
    \toprule
        \multicolumn{2}{c|}{} & \multicolumn{4}{c}{Original PGD} & \multicolumn{4}{c}{Partial-PGD}\\ \midrule
        \multicolumn{2}{c|}{Metrics} & $D_{tr}$ & $D_{tf}$ & Time (s) & US  & $D_{tr}$ & $D_{tf}$ & Time (s) & US \\ \midrule \midrule
        \multirow{4}{*}{CIFAR-10} & VGG16 & 92.03 & 0 & 14.18 &0.9394  & 92.18 & 0 & 3.76 &0.9405 \\ 
        ~ & ResNet18 & 92.97 & 0 & 18.19 &0.9463  & 93.53 & 0 & 4.37 &0.9504 \\ 
        ~ & ResNet50 & 91.84 & 0 & 44.15 &0.9380  & 93.52 & 0 & 7.76 &0.9503\\ 
        ~ & ViT & 78.07 & 0 & 237.36 &0.8442  & 81.14 & 0 & 25.93 &0.8640\\ \midrule
        \multirow{4}{*}{Fashion-MNIST} & VGG16 & 94.15 & 0 & 16.61 &0.9551 & 93.89 & 0 & 8.75 &0.9531\\ 
        ~ & ResNet18 & 94.49 & 0 & 21.35 &0.9576 & 94.54 & 0 & 5.194 &0.9579\\ 
        ~ & ResNet50 & 94.47 & 0 & 51.74 &0.9574 & 94.48 & 0 & 9.14 &0.9575\\ 
        ~ & ViT & 87.4 & 0 & 23.99 &0.9063 & 87.44 & 0 & 13.396 &0.9066\\ \midrule
        \multirow{4}{*}{VGGFace2} & VGG16 & 96.29 & 0 & 19.95 &0.9349 & 96.70 & 0 & 5.60 &0.9743\\ 
        ~ & ResNet18 & 91.42 & 0 & 29.21 &0.9467 & 95.34 & 0 & 6.51 &0.9639\\
        ~ & ResNet50 & 93.02 & 0 & 298.15 &0.9712 & 93.28 & 0 & 17.77 &0.9485\\ 
          & ViT  & 95.76  & 0  & 18.65 &0.9672  & 95.5  &0   & 6.748 &0.9651
  \\ \bottomrule
    \end{tabular}
    }
    
\end{table}
% \usepackage{multirow}
% \usepackage{graphicx}
\begin{table}[t]
\centering
\caption{Effect of Softmax vs. Double Softmax for all datasets.}
\label{tab:sub_softamx}
\resizebox{\columnwidth}{!}{%
\begin{tabular}{c|c|cccc|cccc} \toprule
                               \multicolumn{2}{c|}{}          & \multicolumn{4}{c|}{w/o Double Softmax} & \multicolumn{4}{c}{w/ Double Softmax} \\ \midrule
                               \multicolumn{2}{c|}{Metrics}         & $D_{tr}$    & $D_{tf}$    & Time (s) &US  & $D_{tr}$    & $D_{tf}$   & Time (s)   &US\\ \midrule \midrule 
\multirow{4}{*}{CIFAR-10}      & VGG16    & 92.02       & 0           & 3.88   &0.9472    & 92.18       & 0          & 3.76      &0.9405 \\
                               & ResNet18 & 93.10       & 0           & 4.46   &0.9430    & 93.53       & 0          & 4.37      &0.9504 \\
                               & ResNet50 & 92.53       & 0           & 7.56   &0.9393    & 93.52       & 0          & 7.76      &0.9503 \\
                               & ViT      & 78.73       & 0           & 69.42  &0.8484    & 81.14       & 0          & 25.93     &0.8640 \\ \midrule
\multirow{4}{*}{Fashion-MNIST} & VGG16    & 84.74       & 0           & 10.90  &0.8880    & 93.89       & 0          & 8.75      &0.9531 \\
                               & ResNet18 & 91.42       & 0.1         & 25.87  &0.9341    & 94.54       & 0          & 5.19      &0.9579 \\
                               & ResNet50 & 80.91       & 0           & 93.49  &0.8625    & 94.48       & 0          & 9.13      &0.9575 \\
                               & ViT      & 87.01       & 0           & 61.37  &0.9036    & 87.44       & 0          & 13.39 &0.9066     \\ \midrule
\multirow{4}{*}{VGGFace2}     & VGG16    & 92.94       & 0           & 3.71    &0.9505   & 96.70       & 0          & 5.60       &0.9743 \\
                               & ResNet18 & 93.54       & 0           & 8.75   &0.9468    & 95.34       & 0          & 6.51      &0.9639 \\
                               & ResNet50 & 93.03       & 0           & 26.90  &0.9461    & 93.28       & 0          & 17.77      &0.9485 \\
                               & ViT      & 94.91       & 0           & 8.49   &0.9608    & 95.50       & 0          & 6.74   &0.9651  \\  \bottomrule
\end{tabular}%
}
\end{table}

Our final US is constructed and derived to Eq.~\ref{eq:ue_2} as follows:
\begin{align}
\text{US} (\text{acc}_{r},\text{acc}_{f}) & = \cfrac{\cfrac{\text{US}_r + \text{US}_f}{2}-\text{exp}(0)}{\text{exp}(1)-\text{exp}(0)}\label{eq:12}
\\ &=  \cfrac{\text{US}_r + \text{US}_f - 2 \cdot \exp(0)}{2 \cdot (\exp(1)-\exp(0))} \label{eq:13}
\\ & = \cfrac{\exp(\cfrac{\text{acc}_r}{100})+\exp(1-\cfrac{\text{acc}_f}{100})-2}{2 \cdot (\exp(1)-1)} \notag
\end{align}
By introducing this novel metric, US, we can more effectively characterize and evaluate whether an unlearning method has properly forgotten information from forgetting data, while retained information from retain data simultaneously. Throughout experiments, we show that US effectively characterize and capture the underlying performance of forgetting and retain data performance across different proposed methods.

\begin{comment}
    우리는 unlearning의 performance가 forgetting data에 대한 정확도는 낮아야하고, retain data에 대한 정확도는 높아야한다는 사실을 인지하고 있다. 예를 들어, 한 unlearning method가 forgetting data에 대한 낮은 정확도를 가지나 retain해야할 데이터의 정보를 잘 보존하지 못한 경우와, 다른 unlearning method가 retain해야할 정보는 잘 보존하나, forgetting data를 잘 잊지 못하는 경우, 두 경우,가 있다고 생각할 수 있다. 우리는 forgetting data 혹은 retain data 하나에만 초점을 두고 평가하지 않기 때문에 어떤 방법이 더 좋은지 쉽게 결정할 수 없다. 그러므로 이 두 정확도에 대한 평가를 동시에 고려하는 지표가 존재해야한다.
    우리는 두 값의 단순한 차로 unlearning method의 평가를 내리기에는 조금 부적절하다고 생각했다. 그래서, 지수함수를 사용하여, 각 값의 가중치를 동시에 고려한 새로운 metric을 정의했다. 다시 말해, 좋은 수치일 수록 좋은 값을 가지고, 그렇지 못할 수록 낮은 값을 가지게 한다. 우리의 정확도는 0부터 100이하의 값으로 표기하였기 때문에 먼저 100으로 나누어 0과 1사이의 값을 가지게 한다. 이 값을 지수함수에 넣게 되는데, 이때 forgetting data 파트는 x축 방향으로 1만큼 평행이동을 한다. 계산된 두 값을 exp(1)과 exp(0)으로 min-max scaling(normalizing)하고 둘의 평균을 우리는 Unlearning Score라고 정의하였다. 그리하여 US가 1에 가까울 수록 좋은 unlearning method이고, 0에 가까울 수록 안좋은 method라고 할 수 있다.
    Good unlearning performane means that the model forget information of the forgetting data and simultaneously retain the information of the retain data. On other words, it must have low in forgetting data accuray and high in retain data. But, it is hard to measure what is the more efficient unlearning method because of a reference point. For example, if one unlearning method shows low forgetting data accuracy but bad performane for retain data, and another is the opposite, bad performance for forgetting data and high retain data accuracy, we can confuse to grade the methods.
\end{comment}

% \input{main/supple_table/total_alpha}
\section{Hyper-parameters effects in KD}
\begin{comment}
    Table 9는 knowledge distillation의 hyper-parameter인 \alpha의 변화량에 따른 unlearning performance를 나타낸 것이다. 만약 alpha가 0이라면, 우리의 loss function에서 $\machcal{L}_{CE}$만을 사용하는 것으로, 이는 오직 forgetting data를 forget하는데만 초점이 맞춰져있다. 그렇기 때문에 boundary의 정보를 잘 보존하지 못하여 최대 40퍼센트 정도가량 하락할 때도 있다.
만약 alpha가 1이라면, $\machcal{L}_{CU}$만을 사용하는 것으로, 이는 boundary를 보존하는데 초점이 맞춰져 있다. 그렇기 때문에 boundary의 정보를 잘 담고있다고 해도, forgetting data를 제대로 잊지 못하는 경우가 발생한다.
그러므로 적절한 \alpha를 통해 Knowledge distillation으로 forgetting data와 boundary 둘 사이의 균형을 유지하는 것이 중요하다.
\end{comment}
Table~\ref{tab:total_alpha-table} illustrates the variations in unlearning performance based on the hyper-parameter $\alpha$ in knowledge distillation. When $\alpha$ is set to 0, our loss function $\mathcal{L}$ employs only $\mathcal{L}_{CE}$, focusing solely on forgetting data. As a result, it may not effectively retain information from the boundary, leading to a potential drop of up to approximately 40\%. On the other hand, setting $\alpha$ to 1 utilizes only $\mathcal{L}_{DI}$, prioritizing the retention of the boundary. Although this approach may preserve boundary information well, it might struggle to forget the forgetting data properly. Hence, striking the right balance between forgetting data and boundary information through an appropriate $\alpha$ value in knowledge distillation is crucial, as shown in Table~\ref{tab:total_alpha-table}.
Table~\ref{tab:KD_T} illustrates the variation in unlearning performance based on the hyper-parameter $T$ in knowledge distillation. In our experiments, $T$ = 4 yielded the best performance; however, variations in 
$T$ showed a difference in accuracy of 0.2\% as indicated in Table~\ref{tab:KD_T} under $D_{tr}$. Table~\ref{tab:KD_2T} illustrates the variation in unlearning performance based on the exponent of $T^{x}$ in Eq.~\ref{eq:final_loss}, $T$ is fixed at 4. In our experiments, $x$ = 2 yielded the best performance, whereas values greater than 3 demonstrated poorer performance. Experiments with hyper-parameters tuning show that appropriately selecting values in knowledge distillation can yield the better performance in the unlearning task.
% 다음으로 하이퍼 파라미터 T의 영향을 확인하기 위해 Table.7 에서 T를 변경하여 unlearning 성능을 ResNet18 Cifar-10에서 확인하였다. 우리의 실험에서는 T를 4로 설정하였으며 가장 좋은 성능이 나왔지만 T의 변화의 차이는 Table 7의 Dtr에서 처럼 0.2% 정확도 차이를 보여주었다. Table 8은 Eq.7내 T의 값을 4로 고정하고 T^x의 x값을 변경했을 때의 unlearning performance를 측정하였다. 우리의 실험에서 x=2일 때 좋은 성능을 보여 주었고 3 이상에서는 성능이 좋지 않음을 보여주었다. 하이퍼 파라미터 변경 실험을 통해 지식 증류에서의 값을 적절하게 취하면 unlearning task에서 좋은 성능을 보여줄 것 있다.

% \input{main/supple_table/total_alpha}

\section{Original PGD vs. Partial-PGD}
% \input{main/supple_table/sup_pgd_vs_partial_pgd}
\begin{comment}
    우리는 Partial-PGD의 시간적 효율성과, 성능적인 우위를 증명하기위해 다양한 모델과 데이터셋에서 실험을 진행하였다. Table(새로운 테이블)에서, original PGD도 unlearning에 있어 뛰어난 성능을 보인다. 그러나 Partial-PGD는 original PGD보다 같거나 근소적으로 높은 성능을 보이며 특히 ResNet50의 VGGFace2에서는 최대 16.77배 시간을 절약할 수 있다. Original PGD는 모델의 전체 구조를 이용하기 때문에 모델이 크고 깊을수록 더 많은 시간 소모를 요구한다. 반면에 Fig.1에서 표현된 것 처럼 Partial-PGD는 일부 layer만 이용해서 원하는 목적을 빠른 시간내에 달성할 수 있기 때문에 더욱 효과적이라고 할 수 있다.
\end{comment}
We conduct experiments on various models and datasets to demonstrate the temporal efficiency and performance advantage of Partial-PGD. In Table~\ref{tab:sub_pgd}, the original PGD also presents an excellent performance in terms of unlearning. However, we show that Partial-PGD exhibits comparable or superior performance to original PGD, notably in VGGFace2 with ResNet50. On the other hand, it can save unlearning process time up to 16.77 times. The original PGD requires more time as the model has to utilize the complete model layers. In contrast, as depicted in Fig.~\ref{fig:pgdvsppgd}, Partial-PGD can be considered more effective, as it only uses particular layers to achieve the desired objectives faster.

\section{Effectiveness of Double Softmax}
% \input{main/supple_table/sup_soft_vs_d_soft}
\begin{comment}
    Eq.5와 같이 double softmax를 사용해 다양한 데이터셋과 모델에 성능에 대한 강건함을 주었다. Table.7에서는 데이터 세트별, 모델 별 double softmax의 효과를 확인하고자 실험을 실시했다. 전체적으로 double softmax가 unlearning 수렴 속도가 빨랐다. 또한 근소한 차이지만 실험적으로 대부분의 모델에서 정확도 측면으로 더 좋은 성능을 보였다. 특히 Fashion-MNIST에서 두드러진 수치를 확인할 수 있다. double softmax가 더 soft한 logits을 생성하여 adversarial examples에 대한 이상치에 대해 강건해지며 훈련 안정성이 올라간 것으로 보여진다.
\end{comment}
As shown in Eq.~\ref{eq:double_softmax}, double Softmax provides performance robustness across various datasets and models. In Table~\ref{tab:sub_softamx}, we conduct experiments to examine the effects of double Softmax across different datasets and models. Overall, double Softmax facilitates a faster unlearning convergence speed. Furthermore, though the difference is marginal, our experimental results demonstrate higher accuracy performance across most models. Especially in the case of Fashion-MNIST, notable improvements can be observed. Double Softmax generates softer logits, enhancing robustness against outliers of adversarial examples and improving training stability.

\section{Additional Ablation on Data Usage Ratio}
% \usepackage{multirow}
% \usepackage{graphicx}
\begin{table*}[t]
\centering
\caption{Unlearning performance with varying amounts of data used for unlearning.}
\label{tab:total_amount}
\resizebox{\textwidth}{!}{%
\begin{tabular}{c|c|ccc|ccc|ccc|ccc} \toprule
                                \multicolumn{2}{c|}{Model}                                                           & \multicolumn{3}{c|}{VGG16} & \multicolumn{3}{c|}{ResNet18} & \multicolumn{3}{c|}{ResNet50} & \multicolumn{3}{c}{ViT} \\ \midrule
                               \multicolumn{2}{c|}{Total Extra Data Used} & 100\%    & 50\%    & 10\%    & 100\%     & 50\%     & 10\%     & 100\%     & 50\%     & 10\%     & 100\%   & 50\%    & 10\%   \\ \midrule \midrule
\multirow{4}{*}{CIFAR-10}      & $D_{tr}$                                                        & 92.18   & 92.42  & 92.38  & 93.53    & 93.51   & 93.38   & 93.52    & 93.63   & 93.37   & 81.14  & 81.14  & 81.60 \\
                               & $D_{tf}$                                                        & 0       & 0      & 0      & 0        & 0       & 0       & 0        & 0       & 0       & 0      & 0      & 0     \\
                               & Time                                                            & 3.76    & 1.91   & 1.21   & 4.37     & 2.28    & 1.45    & 7.76     & 3.81    & 1.62    & 25.93  & 25.63  & 14.55 \\ 
                               & US & 0.9405    & 0.9422   & 0.9420   & 0.9504     & 0.9503    & 0.9493    & 0.9503     & 0.9512    & 0.9493    & 0.8640  & 0.8640  & 0.8662\\ \midrule
\multirow{4}{*}{Fashion-MNIST} & $D_{tr}$                                                        & 93.89   & 94.23  & 93.74  & 94.54    & 94.67   & 97.19   & 94.48    & 94.21   & 84.88   & 87.44  & 87.09  & 87.46 \\
                               & $D_{tf}$                                                        & 0       & 0      & 0      & 0        & 0       & 0       & 0        & 0       & 0       & 0      & 0      & 0.1   \\
                               & Time                                                            & 8.75    & 2.20   & 0.48   & 5.19     & 1.96    & 0.63    & 9.14     & 4.54    & 1.04    & 13.39  & 4.88   & 2.69  \\ 
                               & US & 0.9531    & 0.9556   & 0.9520   &0.9579     & 0.9589    & 0.9487    & 0.9575     & 0.9555    & 0.8890    & 0.9066  & 0.9042   & 0.9060 \\ \midrule
\multirow{4}{*}{VGGFace2}      & $D_{tr}$                                                        & 96.70   & 95.83  & 95.88  & 95.34    & 94.35   & 94.46   & 93.28    & 94.24   & 93.13   & 95.50  & 95.82  & 95.88 \\
                               & $D_{tf}$                                                        & 0       & 0      & 0      & 0        & 0       & 0       & 0        & 0       & 0       & 0      & 0      & 0     \\
                               & Time                                                            & 5.60    & 5.12   & 5.36   & 6.51     & 4.22    & 1.8     & 17.77    & 23.09   & 15.35   & 6.74   & 2.46   & 2.04 \\
                               & US  & 0.9743    & 0.9677   & 0.9681   & 0.9639     & 0.9565    & 0.9573     & 0.9485    &0.9557   & 0.9475   & 0.9651   & 0.9676   & 0.9680\\ \bottomrule
\end{tabular}%
}
\end{table*}
\begin{comment}
우리의 알고리즘에서 unlearning process에 투입되는 임의의 D_f를 선택하여 양을 줄여도 unlearning이 가능하지 성능에 영향이 없는 지에 대한 Ablation Study를 진행한다. Table.8 에서 추가적으로 다양한 데이터 셋에서 unlearning process에 투입되는 데이터 사용량을 줄였을 때 결과를 나타낸다. 놀라운 사실은 D_f의 양을 줄여도 정확도 측면에서 성능이 뒤지지 않는다는 사실이고 unlearning process를 완료 시간도 줄어든 것을 볼 수 있다. ViT의 fashion-MNIST에는 D_rf가 0.1% 남아 있는 것을 볼 수 있다.
\end{comment}
We conduct an Ablation Study to investigate whether reducing the amount of randomly selected forgetting data involved in our algorithm's unlearning process impacts performance while maintaining the possibility of unlearning. Table~\ref{tab:total_amount} presents the results when reducing the data used in the unlearning process across various datasets. The remarkable finding is that even with a reduction in the quantity of forgetting data, there is no significant decline in performance from an accuracy perspective. Additionally, a decrease in the completion time of the unlearning process can also be observed. It can be observed that for ViT on Fashion-MNIST, the accuracy of $D_{tf}$ remains at 0.1\%.

% \usepackage{graphicx}
\begin{table*}[h]
\centering
\caption{Unlearning performance for each class on CIFAR-10}
\label{tab:multi_cifar10}
\resizebox{0.8\textwidth}{!}{%
\begin{tabular}{c|c|cccccccccc} \toprule
         % & \multicolumn{10}{c}{Forget class}                                             \\ \midrule
          \multicolumn{2}{c|}{Forgetting Class} & 0     & 1     & 2     & 3     & 4     & 5     & 6     & 7     & 8     & 9     \\ \midrule \midrule
\parbox[t]{2mm}{\multirow{6}{*}{\rotatebox[origin=c]{90}{VGG16}}}&$D_{r}$  & 99.98 & 99.98 & 99.98 & 99.98 & 99.98 & 99.98 & 99.98 & 99.98 & 99.97 & 99.97 \\
&$D_{f}$  & 0     & 0     & 0     & 0     & 0     & 0     & 0     & 0     & 0     & 0     \\
&$D_{tr}$ & 92.7  & 92.2  & 93.35 & 94.44 & 93.17 & 93.98 & 93.54 & 92.47 & 92.24 & 92.35 \\
&$D_{tf}$ & 0     & 0     & 0     & 0     & 0     & 0     & 0     & 0     & 0     & 0     \\
&Time (s) & 3.75  & 7.31  & 3.7   & 3.71  & 3.68  & 3.66  & 2.5   & 3.68  & 3.65  & 3.72 \\ 
&US &0.9443	&0.9406	&0.9491	&0.9572	&0.9477	&0.9537	&0.9431	&0.9426	&0.9409	&0.9417 \\ \midrule
          
\parbox[t]{2mm}{\multirow{6}{*}{\rotatebox[origin=c]{90}{ResNet18}}}&$D_{r}$  & 99.97 & 99.98 & 99.98 & 99.98 & 99.97 & 99.97 & 99.98 & 99.98 & 99.98 & 99.98 \\
&$D_{f}$  & 0     & 0     & 0     & 0     & 0     & 0     & 0     & 0     & 0     & 0     \\
&$D_{tr}$ & 93.84 & 93.44 & 94.39 & 95.26 & 93.86 & 94.53 & 93.53 & 93.48 & 93.58 & 93.51 \\
&$D_{tf}$ & 0     & 0     & 0     & 0     & 0     & 0     & 0     & 0     & 0     & 0     \\
&Time (s) & 4.37  & 4.42  & 4.44  & 4.41  & 3.52  & 3.25  & 3.32  & 4.40  & 4.42  & 4.47  \\ 
&US &0.9527	&0.9497	&0.9568	&0.9633	&0.9528	&0.9578	&0.9504	&0.9500	&0.9508	&0.9502 \\ \midrule

\parbox[t]{2mm}{\multirow{6}{*}{\rotatebox[origin=c]{90}{ResNet50}}}&$D_{r}$  & 99.94 & 99.93 & 99.94 & 99.94 & 99.94 & 99.97 & 99.94 & 99.92 & 99.93 & 99.89 \\
&$D_{f}$  & 0     & 0     & 0     & 0     & 0     & 0     & 0     & 0     & 0     & 0     \\
&$D_{tr}$ & 93.93 & 93.07 & 94.45 & 95.26 & 94.06 & 94.54 & 93.56 & 93.48 & 93.58 & 92.97 \\
&$D_{tf}$ & 0     & 0     & 0     & 0     & 0     & 0     & 0     & 0     & 0     & 0     \\
&Time (s) & 7.42  & 7.41  & 7.52  & 7.60  & 7.49  & 7.86  & 7.49  & 7.54  & 7.47  & 7.47  \\ 
&US &0.9534	&0.9470	&0.9572	&0.9633	&0.9543	&0.9579	&0.9506	&0.9500	&0.9508	&0.9463\\ \midrule

\parbox[t]{2mm}{\multirow{6}{*}{\rotatebox[origin=c]{90}{ViT}}}&$D_{r}$  & 88.43 & 88.2  & 88.42  & 89.74 & 87.83 & 88.96 & 86.48  & 86.72 & 87.52 & 88.07 \\
&$D_{f}$  & 0     & 0     & 0.02   & 0     & 0     & 0     & 0      & 0     & 0     & 0     \\
&$D_{tr}$ & 82.68 & 81.60 & 83.31  & 84.10 & 82.14 & 82.65 & 80.73  & 80.84 & 81.13 & 82.07 \\
&$D_{tf}$ & 0     & 0     & 0      & 0     & 0     & 0     & 0      & 0     & 0     & 0     \\
&Time (s) & 15.20 & 16.68 & 250.82 & 21.16 & 46.01 & 33.49 & 104.40 & 33.68 & 25.20 & 8.65 \\ 
&US &0.8742	&0.8670	&0.8776	&0.8837	&0.8706	&0.8732	&0.8613	&0.8620	&0.8639	&0.8701 \\ \bottomrule

\end{tabular}%
}
\end{table*}
\section{Unlearning Performance on Every Class}
\begin{comment}
    '우리는 모든 클래스에 대해 우리의 방법을 실험했다. 우리는 데이터셋과 클래스를 가리지 않고 좋은 성능임을 보여준다. Table 10은 CIFAR-10에 대한 실험으로서, 우리의 방법이 최소 2.5초 만에 어떤 한 클래스 전체를 지우면서도 나머지 정보는 유지하는 성능을 보여준다. Table 11은 Fahsion-MNIST에 대한 실험으로서, 완벽하게 한 클래스를 지우지 못하는 경우도 있지만, 우리의 방법은 다른 모든 실험에 대해 효율적이고 효과적인 성능임을 보여주고 있다. 마지막으로 Table 12는 VGGFace2에 대한 실험으로서, 우리의 실험은 클래스별 유사성이 뛰어난 얼굴 데이터셋에 대해서도 아주 뛰어난 성능임을 보여준다.'
\end{comment}
% \usepackage{graphicx}
\begin{table*}[ht!]
\centering
\caption{
Unlearning performance for each class on Fashion-MNIST}
\label{tab:multi_fashion}
\resizebox{0.8\textwidth}{!}{%
\begin{tabular}{c|c|cccccccccc} \toprule
         \multicolumn{2}{c}{Forgetting Class} & 0     & 1     & 2     & 3     & 4     & 5     & 6     & 7     & 8     & 9     \\ \midrule \midrule

\parbox[t]{2mm}{\multirow{6}{*}{\rotatebox[origin=c]{90}{VGG16}}}&$D_{r}$  & 99.84	&99.75	&99.74	&99.62	&99.84	&99.2	&99.88	&99.88	&99.78	&99.85 \\
&$D_{f}$  & 0     & 0     & 0     & 0     & 0     & 0     & 0     & 0     & 0     & 0     \\
&$D_{tr}$ & 96.02	&94.1	&95.27	&94.93	&95.37	&93.53	&97.3	&94.83	&94.31	&94.5 \\
&$D_{tf}$ & 0     & 0     & 0     & 0     & 0     & 0     & 0     & 0     & 0     & 0     \\
&Time (s) & 4.35	&8.56	&4.24	&4.32	&4.3	&4.36	&4.33	&4.36	&4.35	&4.33 \\ 
&US &0.9691	&0.9546	&0.9634	&0.9608	&0.96422	&0.9504	&0.9789	&0.9601	&0.9562	&0.9576 \\ \midrule

\parbox[t]{2mm}{\multirow{6}{*}{\rotatebox[origin=c]{90}{ResNet18}}}&$D_{r}$  & 97.44	&97.21	&97.94	&97.65	&96.35	&97.05	&98.72	&98.31	&97.9	&98.41  \\
&$D_{f}$  & 0     & 0     & 0     & 0     & 0     & 0     & 0     & 0     & 0     & 0     \\
&$D_{tr}$ & 94.94	&93.75	&95.36	&94.63	&93.84	&93.48	&96.71	&94.94	&94.48	&95.08 \\
&$D_{tf}$ & 0     & 0     & 0     & 0     & 0     & 0     & 0     & 0     & 0     & 0     \\
&Time (s) & 7.77	&15.87	&53.49	&25.62	&15.44	&20.89	&15.32	&10.24	&5.15	&5.22 \\ 
&US & 0.9609	&0.9520	&0.96415	&0.9578	&0.9527	&0.9500	&0.9743	&0.9609	&0.9575	&0.9620 \\ \midrule
\parbox[t]{2mm}{\multirow{6}{*}{\rotatebox[origin=c]{90}{ResNet50}}}&$D_{r}$  & 97.47  & 98.02  & 96.45 & 98.05 & 97.22 & 98.16 & 98.35 & 98.13 & 98.20 & 98.27 \\
&$D_{f}$  & 0.05   & 0      & 0     & 0     & 0     & 0     & 0     & 0     & 0     & 0     \\
&$D_{tr}$ & 94.94  & 94.26  & 94.11 & 95.23 & 94.76 & 94.53 & 96.37 & 94.62 & 94.63 & 95.05 \\
&$D_{tf}$ & 0      & 0      & 0     & 0.1   & 0     & 0     & 0     & 0     & 0     & 0     \\
&Time (s) & 105.24 & 118.87 & 99.68 & 9.14  & 99.77 & 8.83  & 25.99 & 17.50 & 10.86 & 9.33 \\ 
&US &0.9609	&0.9558	&0.9547	&0.9631	&0.9596	&0.9578	&0.9718	&0.9585	&0.9586	&0.9617 \\ \midrule

\parbox[t]{2mm}{\multirow{6}{*}{\rotatebox[origin=c]{90}{ViT}}}&$D_{r}$  & 93.03 & 90.52 & 91.02 & 91.72 & 92.82 & 89.82 & 93.22 & 91.99 & 89.82 & 91.63 \\
&$D_{f}$  & 0     & 0     & 0     & 0     & 0     & 0     & 0     & 0     & 0     & 0     \\
&$D_{tr}$ & 90.37 & 87.63 & 88.72 & 89.23 & 90.74 & 86.91 & 91.45 & 88.93 & 87.17 & 88.54 \\
&$D_{tf}$ & 0     & 0     & 0     & 0     & 0     & 0     & 0     & 0     & 0     & 0     \\
&Time (s) & 9.88  & 4.93  & 19.8  & 9.77  & 19.7  & 4.95  & 14.71 & 4.93  & 9.82  & 4.91  \\
&US &0.9273	&0.9079	&0.9156	&0.9192	&0.9300	&0.9029	&0.9351	&0.9171	&0.9047	&0.9143 \\ \bottomrule
\end{tabular}%
}
\end{table*}
% \usepackage{graphicx}
\begin{table*}[ht]
\centering
\caption{Unlearning performance for each class on VGGFace2}
\label{tab:multi_face}
\resizebox{0.8\textwidth}{!}{%
\begin{tabular}{c|c|cccccccccc} \toprule
         \multicolumn{2}{c|}{Forgetting Class}& 0     & 1     & 2     & 3     & 4     & 5     & 6     & 7     & 8     & 9     \\ \midrule \midrule
\parbox[t]{2mm}{\multirow{6}{*}{\rotatebox[origin=c]{90}{VGG16}}}&$D_{r}$  & 99.71 & 99.89 & 99.78 & 99.59 & 99.52 & 99.38 & 99.87 & 99.74 & 99.44 & 99.73 \\
&$D_{f}$  & 0     & 0     & 0     & 0     & 0     & 0     & 0     & 0     & 0     & 0     \\
&$D_{tr}$ & 96.01 & 97.11 & 96.39 & 96.34 & 95.87 & 96.03 & 97.12 & 96.25 & 95.72 & 96.82 \\
&$D_{tf}$ & 0     & 0     & 0     & 0     & 0     & 0     & 0     & 0     & 0     & 0     \\
&Time (s) & 19.27 & 20.01 & 15.25 & 5.92  & 6.35  & 5.74  & 6.45  & 7.63  & 5.86  & 6.42 \\ 
&US &0.9690	&0.9774	&0.9719	&0.9715	&0.9679	&0.9692	&0.9775	&0.9708	&0.9668	&0.9752 \\ \midrule
\parbox[t]{2mm}{\multirow{6}{*}{\rotatebox[origin=c]{90}{ResNet18}}}&$D_{r}$  & 99.85 & 99.85 & 99.60 & 99.82 & 99.68 & 99.79 & 99.54 & 99.83 & 99.72 & 99.94 \\
&$D_{f}$  & 0     & 0     & 0     & 0     & 0     & 0     & 0     & 0     & 0     & 0     \\
&$D_{tr}$ & 94.58 & 94.70 & 94.59 & 95.23 & 95.39 & 95.08 & 95.21 & 95.11 & 95.25 & 95.55 \\
&$D_{tf}$ & 0     & 0     & 0     & 0     & 0     & 0     & 0     & 0     & 0     & 0     \\
&Time (s) & 17.94 & 19.22 & 10.91 & 16.62 & 8.81  & 8.27  & 9.1   & 10.64 & 8.38  & 8.84 \\ 
&US &0.9582	&0.9591	&0.958	&0.9631	&0.9643	&0.9620	&0.9630	&0.9622	&0.9633	&0.9655 \\ \midrule

\parbox[t]{2mm}{\multirow{6}{*}{\rotatebox[origin=c]{90}{ResNet50}}}&$D_{r}$  & 95.19	&98.51	&98.42	&97.84	&98.07	&98.67	&98.62	&98.61	&98.72	&98.61 \\
&$D_{f}$  & 0.05   & 0      & 0     & 0     & 0     & 0     & 0     & 0     & 0     & 0     \\
&$D_{tr}$ & 95.88	&93.9	&93.94	&93.64	&93.49	&94.61	&94.73	&94.46	&94.62	&94.76 \\
&$D_{tf}$ & 0      & 0      & 0     & 0   & 0     & 0     & 0     & 0     & 0     & 0     \\
&Time (s) &22.78	&20.06	&22.17	&17.57	&17.65	&16.84	&18.26	&21.6	&16.92	&18.74
 \\ 
 &US &0.9680	&0.9531	&0.9534	&0.9512	&0.9501	&0.9584	&0.9593	&0.9573	&0.9585	&0.9596 \\ \midrule
\parbox[t]{2mm}{\multirow{6}{*}{\rotatebox[origin=c]{90}{ViT}}}&$D_{r}$  & 94.88 & 94.93 & 95.23 & 94.77 & 95.09 & 94.92 & 95.91 & 95.27 & 95.33 & 95.22 \\
&$D_{f}$  & 0     & 0     & 0     & 0     & 0     & 0     & 0     & 0     & 0     & 0     \\
&$D_{tr}$ & 95.38 & 95.02 & 94.76 & 95.23 & 95.55 & 95.40 & 95.21 & 95.43 & 95.88 & 95.23 \\
&$D_{tf}$ & 0     & 0     & 0     & 0     & 0     & 0     & 0     & 0     & 0     & 0     \\
&Time (s) & 6.19  & 6.23  & 6.93  & 4.94  & 5.01  & 9.99  & 6.02  & 6.03  & 5.31  & 5.72 \\ 
&US &0.9642	&0.9615	&0.9596	&0.9631	&0.9655	&0.9644	&0.9630	&0.9646	&0.96802	&0.9631 \\ \bottomrule

\end{tabular}%
}
\end{table*}

We conduct experiments on our method for all classes to showcase its robust performance regardless of datasets and classes. Table~\ref{tab:multi_cifar10} presents experiments on CIFAR-10, demonstrating our method's ability to quickly erase an entire class, while retaining other information in as little as 2.5 seconds. Table~\ref{tab:multi_fashion} shows experiments on Fashion-MNIST, where although perfect erasure of a single class might not always be achieved, our method consistently demonstrates efficient and effective performance across all other experiments. Finally, Table~\ref{tab:multi_face} highlights experiments on VGGFace2, showing our method's remarkable performance even on face datasets with high inter-class similarity.
\clearpage
\end{document}